\ifdefined\pdfminorversion\pdfminorversion=7\fi
\documentclass{article}
\usepackage[T1]{fontenc}
\usepackage{arxiv}
\usepackage{natbib}
\usepackage{newunicodechar}
\newunicodechar{’}{\textquoteright}
\usepackage{amsmath,amssymb,graphicx,booktabs,longtable,array}
\usepackage{placeins}
\usepackage{needspace}
\usepackage{multirow}
\usepackage{enumitem}
\usepackage[table]{xcolor}

\usepackage{hyperref}
\usepackage{subcaption}
\usepackage{threeparttable}
\usepackage{url}

\input{preprint_layout}
\title{MolHop: Scaffold-Hopping Molecular Design with Agentic Reinforcement Learning}
\renewcommand{\shorttitle}{MolHop}
\author[1,2,\textdagger,\textdaggerdbl]{Yanyi Su}
\author[1,2,\textdagger,\textdaggerdbl]{Chenyan Wang}
\author[3]{Yulei Gong}
\author[2,*]{Yeyun Chen}
\author[1,*]{Xiaohui Yang}
\author[1,3,4,*]{Jun Cheng}
\affil[1]{State Key Laboratory of Physical Chemistry of Solid Surface,\protect\\
College of Chemistry and Chemical Engineering, Xiamen University, Xiamen, China}
\affil[2]{Shanghai Artificial Intelligence Laboratory, Shanghai, China}
\affil[3]{Institute of Artificial Intelligence, Xiamen University, Xiamen, China}
\affil[4]{Laboratory of AI for Electrochemistry (AI4EC), Tan Kah Kee Innovation Laboratory (IKKEM), Xiamen, China}
\newcommand{\preprintAuthorNotes}{%
  \preprintSup{\textdagger}Equal contribution.\quad
  \preprintSup{*}Corresponding authors.\\[2pt]
  \preprintSup{\textdaggerdbl}Work done during internships at Shanghai Artificial Intelligence Laboratory.}

\hypersetup{pdftitle={MolHop: Scaffold-Hopping Molecular Design with Agentic Reinforcement Learning}, pdfauthor={Yanyi Su, Chenyan Wang, Yulei Gong, Yeyun Chen, Xiaohui Yang, Jun Cheng}, pdfsubject={Preprint}}
\begin{document}
\maketitle
\begin{abstract}
Molecular design is often formulated as an optimization problem over existing molecular spaces. However, molecular optimization and the discovery of structurally distinct replacement molecules represent two fundamentally different search problems. LLM-based molecular design agents typically perform the former, conducting local refinement within the neighborhood of existing structures to improve target properties through structural modifications. In contrast, the latter requires exploration of a broader chemical space to discover molecules with entirely different scaffolds while preserving the functional characteristics of reference molecules. We propose MolHop, an agentic framework for cross-scaffold molecular discovery that enables models to navigate and edit within a pre-organized chemical space. MolHop organizes approximately 1.6 million molecular scaffolds from commercial molecules into a coarse-to-fine hierarchical action space. The agent selects new exploration regions through scaffold navigation actions and adjusts candidate properties through molecular editing. The framework maintains historical trajectory memory to support long-horizon search processes. On drug molecule and commercial compound tasks, connecting Grok-4.6 and DeepSeek-Flash-V4.1 to the MolHop framework substantially improves their cross-scaffold joint success rates. We develop MolHop-9B through agentic reinforcement learning, and MolHop-9B outperforms generative baselines including GPT-5.6-sol, Claude-5-Opus, and Gemini-3.1-Pro, while generating candidate molecules with larger scaffold differences from the reference molecules. These results demonstrate that effective molecular discovery beyond the traditional optimization paradigm requires structured chemical and action spaces, enabling agents to explore new structural regions rather than merely refine existing molecules.
\end{abstract}

\begin{figure}[!htbp]
\centering
\renewcommand{\thefigure}{1}

\includegraphics[width=\linewidth]{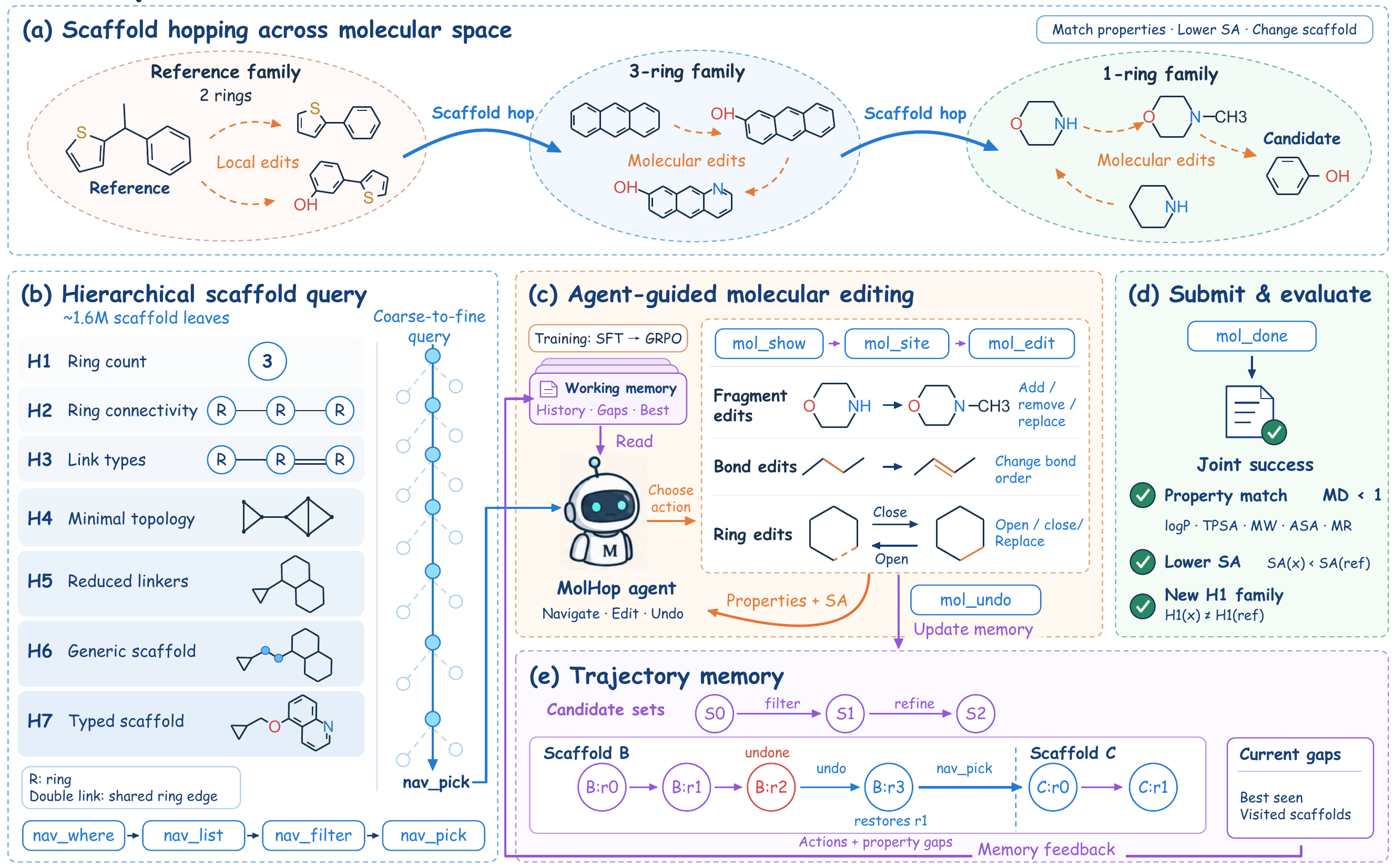}

\caption{Overview of MolHop, a hierarchical scaffold navigation framework with memory-guided molecular editing. (a) Illustration of scaffold hopping across molecular space. (b) Hierarchical scaffold space and coarse-to-fine scaffold query. (c) Agent interaction with MolHop tools for molecular inspection and editing. (d) Joint success criteria for candidate evaluation. (e) Trajectory memory for long-horizon scaffold search.}

\label{fig:1}
\end{figure}

\section{Introduction}

Molecular discovery often starts from a reference compound that already possesses the target function. However, within the molecular discovery process, optimizing an existing molecule and finding a structurally distinct alternative molecule represent two fundamentally different search problems. Molecular optimization typically performs local editing within the neighborhood of the reference structure and optimizes specific properties. In contrast, many practical molecular discovery tasks require candidate molecules to leave the original scaffold family while maintaining multiple target properties. For example, applications including patent circumvention, restricted substance substitution, and toxicity mitigation require alternative molecules with different structures but matched multiple properties. Since a new scaffold starting point may simultaneously alter multiple physicochemical properties, the search process needs to continuously select new structural regions during long-horizon exploration and progressively recover property matching through multiple rounds of scaffold hopping and molecular editing.

In recent years, language model-based molecular design agents have introduced a new paradigm for molecular optimization through tool use. Existing methods \citep{yang2025molact,li2026agentic,li2026reinforcement,zhou2026toolmol} primarily rely on local editing operations to iteratively optimize candidate molecules. Scaffold similarity constraints are further introduced to encourage the search process to preserve the core structure of the reference molecule. In such frameworks, scaffold hopping typically appears only as an incidental outcome of the optimization process. In contrast, cross-scaffold molecular discovery requires active exploration of broader chemical space. Therefore, scaffold hopping should be modeled as a search operation equally important as molecular editing, enabling agents to actively explore new structural regions.

We propose MolHop, an agentic framework for cross-scaffold molecular discovery. As shown in Figure \ref{fig:1}, MolHop models scaffold hopping as a navigation problem in a hierarchical chemical space and integrates scaffold selection, molecular editing, and trajectory memory into a unified agent harness. We extract approximately 1.6 million molecular scaffolds from a commercial molecular library and organize them into a multi-scale action space using a coarse-to-fine hierarchical representation. Through hierarchical navigation actions, the agent can retrieve, filter, and select new exploration regions, and subsequently optimize candidate properties through editing actions. Cross-scaffold search naturally involves multiple rounds of continuous environment interactions. Therefore, MolHop explicitly maintains trajectory memory based on environment feedback to record explored scaffolds, historical actions, and property feedback. This memory serves as an explicit search state, enabling the agent to adjust subsequent navigation and editing strategies based on historical trajectories.

We evaluate MolHop on two tasks involving drug molecules and commercial compounds. Without additional training, directly deploying the MolHop framework with Grok-4.6 and DeepSeek-Flash-V4.1 improves their cross-scaffold joint success rates. Furthermore, we adopt an SFT warm-start followed by GRPO training strategy \citep{guo2025deepseek} and train LLMs with 2B, 4B, and 9B parameter scales using multi-objective optimization rewards. The training objectives include multi-property matching, lower synthetic accessibility (SA) scores \citep{ertl2009estimation}, and satisfying scaffold transition requirements. After agentic reinforcement learning training, MolHop-9B outperforms generative baselines including GPT-5.6-sol, Claude-5-Opus, and Gemini-3.1-Pro on drug molecule and commercial molecule tasks, while discovering candidate molecules with more substantial scaffold differences from the reference molecules. These results demonstrate that combining structured action spaces with search environments that support chemical space navigation can effectively enhance the capability of LLMs for cross-scaffold molecular discovery tasks.

Our main contributions are as follows:

\begin{enumerate}[leftmargin=*]

\item
\textbf{Hierarchical scaffold action space.} We organize approximately 1.6 million commercial molecular scaffolds into a multi-scale space and introduce explicit scaffold navigation actions, enabling agents to perform cross-scaffold search.

\item
\textbf{MolHop harness for long-horizon chemical space exploration.} We integrate scaffold navigation, molecular editing, and trajectory memory into a unified environment to support long-trajectory, multi-turn molecular design processes.

\item
\textbf{Agentic reinforcement learning for cross-scaffold discovery.} We train MolHop agents with multi-objective rewards, enabling them to coordinate scaffold selection and molecular editing under cross-scaffold constraints.

\end{enumerate}

\section{Related Work}

\subsection{Goal-directed molecular search and LLM-based molecular agents}

Goal-directed molecular design has traditionally been formulated as an optimization problem under property feedback. Early approaches optimize molecular generation policies through reinforcement learning or evolutionary search, where candidate molecules are iteratively generated and evaluated according to predefined objectives \citep{olivecrona2017molecular,jensen2019graph}. Subsequent benchmarks and methods improve optimization efficiency through better molecular representations, sampling strategies, and generation policies \citep{brown2019guacamol,gao2022sample,kim2024genetic,lee2025genmol}.

Recent advances in large language models have introduced agentic workflows for molecular design, enabling iterative interaction with molecular editing tools and property evaluation environments. Existing molecular agents typically refine candidate structures through sequential editing actions, where the search process is guided by property feedback \citep{yang2025molact,li2026agentic,zhou2026toolmol}. 

Despite these advances, most existing approaches focus on local structural refinement around an initial molecule. The search trajectory is primarily composed of molecular edits, while transitions between distant scaffold regions are not explicitly modeled as search decisions. However, many practical discovery scenarios, including scaffold replacement and alternative molecule identification, require deliberate exploration of chemically distinct regions while maintaining target properties. MolHop addresses this limitation by introducing scaffold navigation as an explicit action, allowing agents to perform long-range chemical space exploration before local molecular optimization.

\section{Methods}

The MolHop harness comprises three core components, hierarchical scaffold space, a structured molecular action interface, and trajectory memory. As shown in Figure \ref{fig:1}, the agent selects new structural regions in a hierarchical chemical space, adjusts candidate properties through editing actions, and uses feedback from historical trajectories for long-horizon search.

\textbf{Hierarchical Scaffold Space.}
MolHop formulates scaffold hopping as navigation in a hierarchical chemical space. This space organizes scaffolds at multiple levels of structural abstraction, allowing the agent to select structural regions at different granularities according to the search state. We construct the scaffold space from the Enamine Screening Collection\footnote{\url{https://enamine.net/compound-collections/screening-collection}} and obtain approximately 1.6M valid Bemis--Murcko scaffolds. MolHop organizes these scaffolds using the H1-H7 hierarchical scaffold representation, progressively restoring connectivity topology, atom types, and bond types from ring-level structures until reaching concrete editable scaffolds, as shown in Figure \ref{fig:1}(b). This hierarchy provides multiscale scaffold navigation actions, with higher-level nodes supporting exploration across regions and lower-level nodes supporting fine-grained scaffold selection.

In cross-scaffold tasks, MolHop records the H1 region of the reference molecule and checks the region of the candidate structure at the end of the episode to ensure that the final candidate completes the specified scaffold transition. Details of hierarchy construction and the scaffold query process are provided in Figure \ref{fig:A1} and Appendix \ref{sec:app-scaffold-space}.

\textbf{Scaffold Navigation and Molecular Editing Actions.}
MolHop connects scaffold navigation and molecular optimization through a structured action interface. This interface includes four action categories, scaffold navigation, molecular inspection, local editing, and search control, as shown in Table \ref{tab:A2}. The agent selects subsequent actions based on the scaffolds, editable sites, and action constraints returned by the environment, without directly generating molecular strings for structural modification. Scaffold navigation actions act on the hierarchical scaffold space, allowing the agent to filter and select candidate regions according to structural constraints. After selecting a scaffold, the agent uses \texttt{mol\_edit} to perform fragment editing, atom and bond editing, and ring editing, and receives the updated molecular state and property information through environment feedback. This design allows the agent to alternate between scaffold exploration and property optimization within the same exploration trajectory. Detailed action definitions are provided in Appendix \ref{sec:app-actions}.

\textbf{Trajectory Memory.}
Cross-scaffold search requires tracking both explored structural regions and changes in candidate properties across editing steps. MolHop treats memory as a structured search state maintained by the environment and dynamically updates it based on actual interaction events. Memory records the reference targets, current candidate state, property feedback, scaffold visit trajectory, editing history, and search budget, and is returned to the agent together with tool feedback in Figure \ref{fig:1}(e). Because all records originate from actual environment states, property evaluations remain consistent with their corresponding molecular states, and the agent does not need to manually maintain historical structure and property information. This design allows the agent to compare historical candidates, identify explored regions, and adjust subsequent navigation strategies during long-horizon search. Memory supports trajectory decisions within a single episode. Its update rules are controlled by the environment, which produces an agent-facing state according to the context budget. The detailed design is provided in Appendix \ref{sec:app-memory}.

\section{Experiments}

\subsection{Experimental Setup}

\subsubsection{Dataset}

We evaluate the cross-scaffold molecular design ability of MolHop on two tasks involving drug molecules and commercial screening compounds. Given a reference molecule, the model must generate a structurally valid candidate molecule that leaves the reference scaffold family while maintaining five physicochemical properties close to their reference values and lowering the SA score. All methods use the same reference sets and evaluation protocol. Each test set contains 200 reference molecules, drawn respectively from approved synthetic small molecules in ChEMBL \citep{gaulton2012chembl} and the Enamine Screening Collection. Details of dataset construction and splitting are provided in Appendix \ref{sec:app-dataset}.

\subsubsection{Evaluation Metrics}

Given a reference molecule \(x_0\) and a candidate molecule \(x\), we evaluate MolHop in terms of property matching, scaffold transition, synthetic accessibility, and search efficiency. The property matching set \(P\) consists of logP, TPSA, molecular weight MW, Labute ASA, and molar refractivity MolMR. We define: 
\vspace{-0.1cm}
\begin{equation*}
MD(x,x_0)=\sum_{p\in P}
\frac{|p(x)-p(x_0)|}{|p(x_0)|}.
\end{equation*}

We use \(MD<1\) as the property matching condition and require \(SA(x)<SA(x_0)\). Let \(H_k(x)\) denote the level-\(k\) scaffold representation. Joint success is defined as:
\begin{equation*}
JH_k(x,x_0)=
\mathbf{1}[\mathrm{valid}(x)]
\mathbf{1}[MD<1]
\mathbf{1}[SA(x)<SA(x_0)]
\mathbf{1}[H_k(x)\neq H_k(x_0)].
\end{equation*}
\begin{equation*}
\mathrm{joint}@H_k=
\frac{1}{N}\sum_i JH_k(x_i,x_{0,i}).
\end{equation*}

We primarily report joint@H1 and additionally report scaffold Tanimoto similarity and MCS ratio to measure the extent of structural change. 

\subsubsection{SFT and Agentic RL Training}

The MolHop agent uses a two-stage training procedure. We train the 2B, 4B, and 9B models from Qwen3.5 \citep{qwen2026qwen35} to obtain MolHop-2B, MolHop-4B, and MolHop-9B, respectively. First, supervised fine-tuning through SFT teaches the MolHop harness tool-use protocol. The agents are then trained with GRPO in the MolHop environment to learn multi-turn scaffold exploration and molecular optimization policies.

The reinforcement learning stage uses a terminal continuous cost:
\vspace{-0.1cm}

\[
\mathrm{Cost}(x,x_0)
=
MD(x,x_0)
+
\frac{SA(x)}{SA(x_0)},
\qquad
\mathrm{Reward}
=
\max(-\mathrm{Cost},-4).
\]

This continuous reward provides a training signal for property matching and SA optimization, while final performance is evaluated using discrete metrics such as joint@H1. Detailed training configurations, demonstration data construction, and GRPO settings are provided in Appendix \ref{sec:app-training}.

\subsubsection{Baselines and Inference Budget}

We compare three categories of methods, generation baselines, general LLM agents, and MolHop agents trained through agentic RL. Generation baselines receive a reference molecule and a target description, directly generate candidate SMILES, and are then evaluated using a unified offline scoring function. Grok-4.6 and DeepSeek-Flash-V4.1 are connected to the MolHop harness as general LLM agents and perform multi-turn search through tool use and environment feedback without additional training. MolHop agents obtain tool-use policies through the SFT and GRPO procedure described above. We also include the method of \citet{lee2025genmol} as a representative oracle-driven molecular optimization method and optimize it with a budget of 10,000 oracle calls under its original settings. For agent-based methods, the main experiments use a uniform interaction budget of at most 50 assistant turns. Because generation baselines and tool-interaction methods use different search forms, this comparison reflects their final discovery ability under the complete design workflow. We also report the tool execution success rate, average interaction turns, and Efficiency to analyze the interaction costs of different methods.

\begin{table}[!htbp]
\centering

\renewcommand{\arraystretch}{0.75}
\setlength{\tabcolsep}{4pt}

\renewcommand{\thetable}{1}
\caption{Cross-Scaffold Molecular Discovery Performance Comparison}
\label{tab:2}

\begin{tabular}{@{}l l l c c c@{}}
\toprule
\textbf{Dataset} & 
\textbf{Model} & 
\textbf{Method} & 
\textbf{Joint@H1 $\uparrow$} & 
\textbf{Tanimoto median $\downarrow$} & 
\textbf{MCS median $\downarrow$} \\

\midrule

Drug 
& Grok 
& Baseline 
& 41.0\% 
& 0.292 
& 0.913 \\

\rowcolor{gray!15}
& 
& \textbf{With harness} 
& \textbf{64.0\%} 
& \textbf{0.141} 
& \textbf{0.684} \\

& DeepSeek 
& Baseline 
& 15.5\% 
& 0.381 
& 0.966 \\

\rowcolor{gray!15}
& 
& \textbf{With harness} 
& \textbf{52.5\%} 
& \textbf{0.133} 
& \textbf{0.700} \\

\midrule

Commercial 
& Grok 
& Baseline 
& 36.5\% 
& 0.317 
& 0.833 \\

\rowcolor{gray!15}
& 
& \textbf{With harness} 
& \textbf{69.0\%} 
& \textbf{0.145} 
& \textbf{0.600} \\

& DeepSeek 
& Baseline 
& 17.5\% 
& 0.357 
& 0.885 \\

\rowcolor{gray!15}
& 
& \textbf{With harness} 
& \textbf{60.5\%} 
& \textbf{0.132} 
& \textbf{0.600} \\

\bottomrule
\end{tabular}

\end{table}

\subsection{Effect of MolHop Harness on Frozen LLM Agents}

To evaluate the effect of the MolHop harness on the molecular design ability of existing language models, we compare frozen Grok-4.6 and DeepSeek-Flash-V4.1 in one-shot generation mode and with access to the MolHop harness. This comparison changes only the search environment, tool interface, and multi-turn feedback mechanism accessible to the models. As shown in \textbf{Table \ref{tab:2}}, access to the harness significantly improves joint@H1 for both models on the drug and commercial compound tasks. Grok-4.6 improves from 41.0\% and 36.5\% to 64.0\% and 69.0\%, while DeepSeek-Flash-V4.1 improves from 15.5\% and 17.5\% to 52.5\% and 60.5\%. Meanwhile, both scaffold Tanimoto similarity and MCS ratio decrease among successful candidates, indicating that the success rate improvements brought by the harness are accompanied by more substantial structural transitions.

A further decomposition by condition is provided in \textbf{Appendix \ref{sec:app-frozen-agents}}. The analysis shows that the harness improvements do not come solely from improving one constraint, but instead simultaneously increase the proportions satisfying structural validity, property matching, and scaffold transition. This result indicates that a structured search environment can activate the tool-planning ability of frozen language models in cross-scaffold design.

\begin{table}[!htbp]
\centering

\caption{Cross-scaffold molecular discovery results. Joint denotes joint@H1, Tan. denotes scaffold Tanimoto similarity, and MCS denotes scaffold MCS ratio. Higher Joint values indicate better scaffold-constrained molecular discovery performance.}

\label{tab:34}

\scriptsize
\setlength{\tabcolsep}{5pt}
\renewcommand{\arraystretch}{0.75}

\begin{subtable}{0.49\linewidth}
\centering
\caption{Drug dataset}
\label{tab:3}

\begin{tabular}{@{}lccc@{}}
\toprule
Model & Joint@H1 $\uparrow$ & Tanimoto $\downarrow$ & MCS $\downarrow$ \\
\midrule

COPY & 0 & 1.000 & 1.000 \\
ChemLLM-7B & 0.005 & 1.000 & 1.000 \\
Qwen3.5-2B & 0.02 & 1.000 & 1.000 \\
Llama-8B & 0.04 & 0.135 & 0.727 \\
Llama-70B & 0.05 & 0.430 & 0.946 \\
ChemDFM-v2-14B & 0.055 & 1.000 & 1.000 \\
Qwen3.5-4B & 0.055 & 0.467 & 1.000 \\
phi-4 & 0.07 & 0.191 & 0.833 \\
Qwen3.5-9B & 0.075 & 0.227 & 1.000 \\
Qwen3.5-27B & 0.11 & 0.405 & 1.000 \\
DeepSeek-Flash-v4.1 & 0.155 & 0.381 & 0.966 \\
Gemma4-31B & 0.195 & 0.194 & 0.867 \\

\rowcolor{gray!15}
\textbf{MolHop-2B (Ours)} & \textbf{0.285} & 0.108 & 0.632 \\

Grok-4.6 & 0.41 & 0.292 & 0.913 \\
GenMol & 0.415 & 0.103 & 0.889 \\
Gemini-3.1-Pro & 0.435 & 0.291 & 0.938 \\

\rowcolor{gray!15}
\textbf{MolHop-4B (Ours)} & \textbf{0.480} & 0.107 & 0.633 \\

GPT-5.6-Sol & 0.495 & 0.178 & 0.833 \\
Claude-5-Opus & 0.535 & 0.200 & 0.882 \\

\rowcolor{gray!15}
\textbf{MolHop-9B (Ours)} & \textbf{0.555} & 0.123 & 0.660 \\

\bottomrule
\end{tabular}

\end{subtable}
\hfill
\begin{subtable}{0.49\linewidth}
\centering
\caption{Commercial dataset}
\label{tab:4}

\begin{tabular}{@{}lccc@{}}
\toprule
Model & Joint@H1 $\uparrow$ & Tanimoto $\downarrow$ & MCS $\downarrow$ \\
\midrule

COPY & 0 & 1.000 & 1.000 \\
ChemLLM-7B & 0 & 1.000 & 1.000 \\
Qwen3.5-2B & 0.025 & 1.000 & 1.000 \\
Llama-8B & 0.045 & 0.131 & 0.641 \\
ChemDFM-v2-14B & 0.05 & 1.000 & 1.000 \\
Qwen3.5-4B & 0.06 & 0.572 & 1.000 \\
Llama-70B & 0.06 & 0.450 & 0.900 \\
phi-4 & 0.095 & 0.186 & 0.818 \\
Qwen3.5-9B & 0.13 & 0.343 & 0.943 \\
Qwen3.5-27B & 0.165 & 0.468 & 0.964 \\
DeepSeek-Flash-v4.1 & 0.175 & 0.357 & 0.885 \\
Gemma4-31B & 0.225 & 0.244 & 0.778 \\

\rowcolor{gray!15}
\textbf{MolHop-2B (Ours)} & \textbf{0.335} & 0.096 & 0.578 \\

Grok-4.6 & 0.365 & 0.317 & 0.833 \\
Gemini-3.1-Pro & 0.470 & 0.293 & 0.842 \\

\rowcolor{gray!15}
\textbf{MolHop-4B (Ours)} & \textbf{0.525} & 0.102 & 0.568 \\

GenMol & 0.530 & 0.091 & 0.667 \\
Claude-5-Opus & 0.540 & 0.194 & 0.731 \\
GPT-5.6-Sol & 0.615 & 0.185 & 0.737 \\

\rowcolor{gray!15}
\textbf{MolHop-9B (Ours)} & \textbf{0.735} & 0.115 & 0.568 \\

\bottomrule
\end{tabular}

\end{subtable}

\end{table}

\begin{figure}[b]
\centering
\renewcommand{\thefigure}{2}
\includegraphics[width=\linewidth,height=0.70\textheight,keepaspectratio]{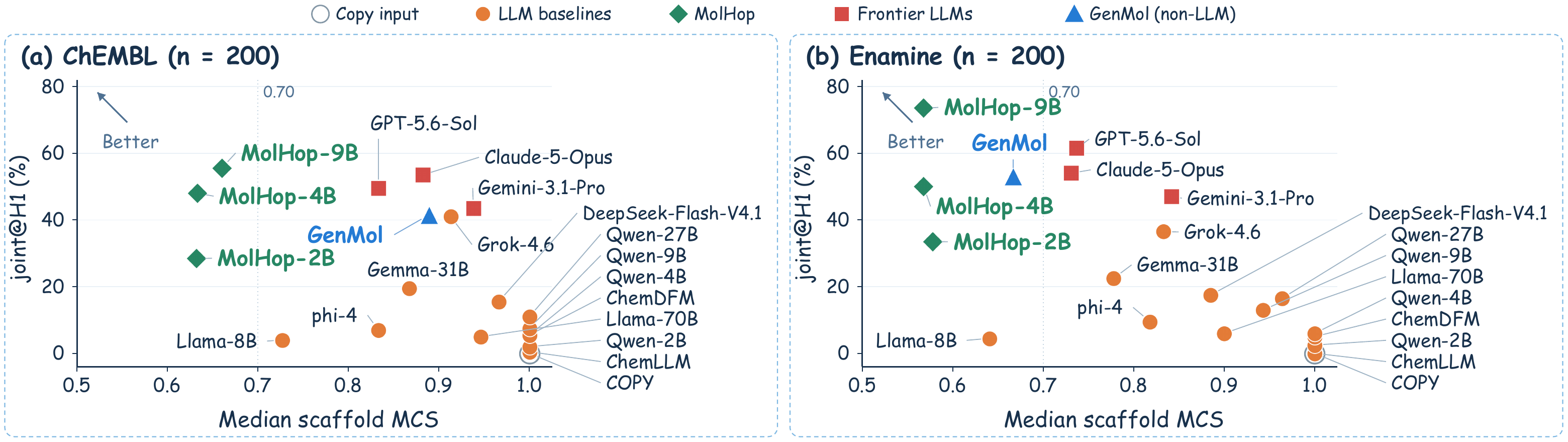}
\caption{(a) Joint@H1 versus median scaffold MCS on the ChEMBL test set. (b) Joint@H1 versus median scaffold MCS on the Enamine test set. Points toward the upper-left region indicate higher joint success with larger scaffold transition.}
\label{fig:2}
\end{figure}

\subsection{Agentic RL Enables Cross-Scaffold Molecular Design}

Within the MolHop harness, we further train agentic policies at three scales, 2B, 4B, and 9B, and evaluate their ability in cross-scaffold molecular design. \textbf{Figure \ref{fig:C1}(a,b)} shows changes in reward and validation joint@H1 during GRPO training. All three model scales obtain higher reward as training progresses and stabilize after approximately 100 optimization steps, while validation joint@H1 also improves with training.

\textbf{Table \ref{tab:34}}(a,b) compare MolHop with generation models and oracle-driven optimization methods on the two test sets. MolHop-9B achieves joint@H1 of 55.5\% and 73.5\% on the ChEMBL and Enamine test sets, respectively, outperforming all one-shot generation baselines and general LLM baselines. Compared with the best generation model, MolHop-9B improves by 2.0 percentage points on the drug task and 12.0 percentage points on the commercial task.

Beyond joint success rate, the scaffold similarity of candidates generated by MolHop also indicates a broader extent of structural transition. MolHop-9B has scaffold Tanimoto medians of 0.123 and 0.115 and MCS medians of 0.660 and 0.568 on ChEMBL and Enamine, respectively. \textbf{Figure \ref{fig:2}(a,b)} further shows the relationship between joint@H1 and scaffold MCS. MolHop models lie closer to the upper-left region of the plots, achieving both higher joint success rates and lower scaffold retention ratios, indicating that the policies learned through agentic RL can coordinate structural transition and property recovery.

We further evaluate joint success under different scaffold transition criteria from H1 to H7, as shown in Figure \ref{fig:C2}. Under more relaxed scaffold criteria, LLM-based generation baselines show an expected increase in joint success. However, this improvement mainly reflects relaxed scaffold constraints rather than stronger scaffold hopping ability, as the scaffold MCS does not substantially decrease.

Increasing model scale further improves the cross-scaffold joint success rate. MolHop-2B, MolHop-4B, and MolHop-9B achieve 28.5\%, 48.0\%, and 55.5\% on ChEMBL and 33.5\%, 52.5\%, and 73.5\% on Enamine, respectively. This improvement primarily reflects the agent's enhanced ability to coordinate scaffold selection, property matching, and SA optimization, rather than simply increasing the magnitude of structural change.

\subsection{Agent behavior under structured action space}

\begin{figure}[t]
\centering
\renewcommand{\thefigure}{3}

\includegraphics[
width=\linewidth
]{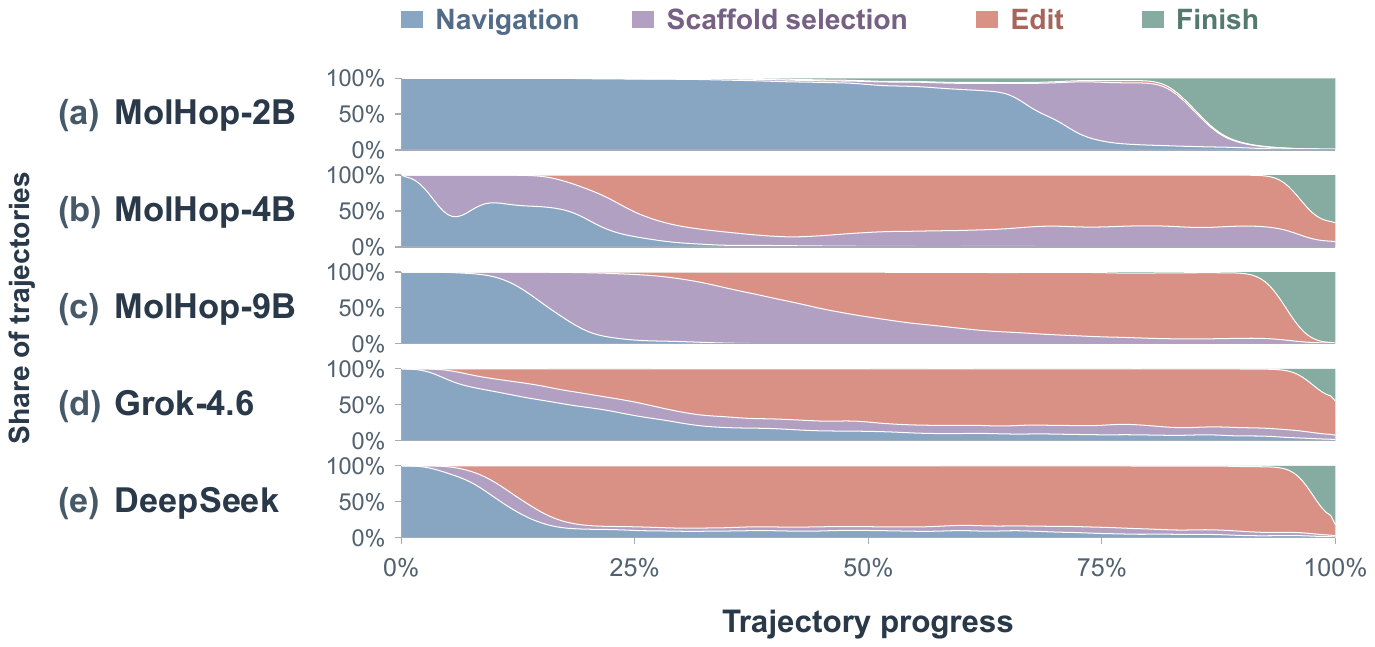}

\vspace{-2mm}

\caption{Tool-use trajectory distributions of MolHop agents and frozen LLM agents. (a) MolHop-2B. (b) MolHop-4B. (c) MolHop-9B. (d) Grok-4.6. (e) DeepSeek-Flash-V4.1. The stacked areas show the fraction of trajectories assigned to the four action categories Navigation, Scaffold Selection, Edit, and Finish over normalized trajectory progress.}

\label{fig:3}
\end{figure}

\textbf{Figure \ref{fig:3}} shows trajectory distributions across four action categories, Navigation, Scaffold Selection, Edit, and Finish, for different models. MolHop agents and general LLM agents exhibit different action allocation patterns. MolHop-2B mainly searches for candidates satisfying the constraints through scaffold navigation and selection and performs relatively few subsequent editing actions. MolHop-4B and MolHop-9B dynamically allocate their search between scaffold selection and editing, switching between exploring new structural regions and adjusting candidate properties. By comparison, general LLM agents typically continue editing-based exploration after the initial scaffold selection, with a smaller proportion of further scaffold changes. Together with the call cost analysis in Appendix \ref{sec:app-tool-efficiency}, this structured action allocation allows MolHop agents to allocate their interaction budgets according to the search state and coordinate scaffold exploration with molecular optimization.

\begin{table}[!htbp]
\centering

\renewcommand{\arraystretch}{0.65}
\setlength{\tabcolsep}{8pt}

\renewcommand{\thetable}{3}

\begin{threeparttable}

\caption{Effect of SFT on Tool-Use Capability}
\label{tab:5}

\begin{tabular}{@{}lccccc@{}}
\toprule
\textbf{Model} & 
\textbf{VER $\uparrow$} & 
\textbf{BOR $\downarrow$} & 
\textbf{EFER $\uparrow$} & 
\textbf{Eff. $\downarrow$} & 
\textbf{ITI $\downarrow$} \\

\midrule

2B Base 
& 21.80\% 
& 71.80\% 
& 0.20\% 
& 1490.9 
& 29.50\% \\

\rowcolor{gray!15}
2B SFT 
& \textbf{91.10\%} 
& \textbf{8.10\%} 
& \textbf{2.20\%} 
& \textbf{441.8} 
& \textbf{0.60\%} \\

\addlinespace[1pt]

4B Base 
& 18.10\% 
& 81.70\% 
& 3.70\% 
& 218.2 
& 3.30\% \\

\rowcolor{gray!15}
4B SFT 
& \textbf{97.80\%} 
& \textbf{2.20\%} 
& \textbf{19.40\%} 
& \textbf{125.1} 
& \textbf{0.40\%} \\

\addlinespace[1pt]

9B Base 
& 4.60\% 
& 95.40\% 
& 13.70\% 
& 174.2 
& 0.90\% \\

\rowcolor{gray!15}
9B SFT 
& \textbf{97.80\%} 
& \textbf{2.00\%} 
& \textbf{19.70\%} 
& \textbf{118.9} 
& \textbf{0.20\%} \\

\bottomrule
\end{tabular}

\begin{tablenotes}
\scriptsize
\item VER denotes Valid Episode Rate. BOR denotes Budget Overflow Rate. EFER denotes Error-free Episode Rate. Eff. denotes Tool-use Efficiency. ITI denotes Invalid Tool Invocation Rate.
\end{tablenotes}

\end{threeparttable}

\end{table}

\vspace{-0.15cm}

\begin{figure}[!htbp]
\centering
\renewcommand{\thefigure}{4}
\includegraphics[width=\linewidth,height=0.70\textheight,keepaspectratio]{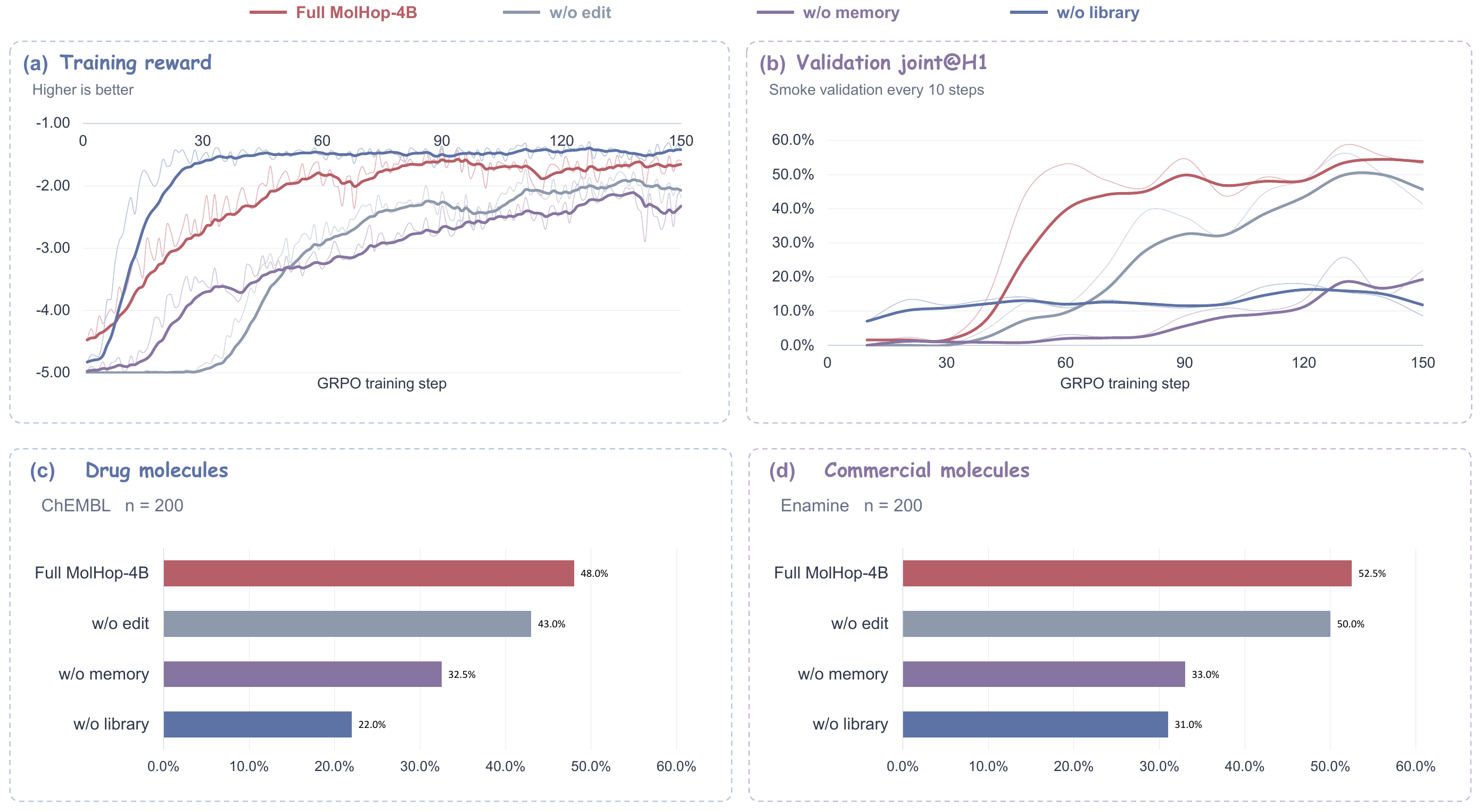}
\caption{Ablation study of MolHop components. Three variants are retrained by removing molecular editing, trajectory memory, or hierarchical scaffold library from MolHop-4B. (a) Training reward. (b) Validation joint@H1. (c) Final joint@H1 on ChEMBL. (d) Final joint@H1 on Enamine.}
\label{fig:4}
\end{figure}

\subsection{Ablation Studies}

\subsubsection{Effect of SFT on Tool-use Ability}

MolHop uses SFT to initialize the agent's tool-use ability and further optimizes its cross-scaffold search policy through RL. To examine the role of SFT, we compare the interaction behavior of base models and models after SFT within the MolHop harness. \textbf{Table \ref{tab:5}} presents tool-use performance at different training stages. SFT significantly improves tool-use reliability across model scales. The Valid Episode Rate VER increases from 21.8\%, 18.1\%, and 4.6\% to 91.1\%, 97.8\%, and 97.8\% for the 2B, 4B, and 9B models, respectively. Meanwhile, both the Budget Overflow Rate BOR and the Invalid Tool Invocation Rate ITI decrease substantially, indicating that SFT effectively establishes the models' understanding of the tool protocol, action parameters, and episode termination procedure.

\subsubsection{Ablation of MolHop Components}

Under the MolHop-4B setting, we separately remove molecular editing, trajectory memory, and the hierarchical scaffold library and retrain the corresponding agents to analyze the contributions of these components to cross-scaffold search.

\textbf{Figure \ref{fig:4}(a-b)} shows the effects of different components on training. The full MolHop maintains favorable optimization trends in both reward and validation joint@H1. Although the variant without the scaffold library obtains higher training reward, its validation joint@H1 decreases substantially. After removing the hierarchical scaffold library, the agent lacks an explicit scaffold navigation path and tends to optimize properties around the current structure. This change affects both scaffold transition and synthetic accessibility optimization. Molecular editing alone makes it difficult to consistently complete the target scaffold transition and to ensure that candidates satisfy the SA improvement condition during property adjustment. The continuous reward can increase through reductions in property error, but optimizing the surrogate objective does not necessarily translate into a higher joint success rate. Detailed analysis is provided in Appendix \ref{sec:app-scaffold-ablation}.

As shown in \textbf{Figure \ref{fig:4}(c,d)}, performance decreases after removing editing, memory, or the scaffold library. Removing the scaffold library and memory leads to more substantial performance decreases, indicating that hierarchical scaffold space and trajectory memory play key roles in long-horizon cross-scaffold search. The improvement from the editing component is relatively small and primarily comes from further adjusting candidate properties after scaffold selection.

% \begin{figure}[!htbp]
% \centering
% \renewcommand{\thefigure}{4}
% \includegraphics[width=\linewidth,height=0.70\textheight,keepaspectratio]{figures/image4.pdf}
% \caption{Ablation study of MolHop components. Three variants are retrained by removing molecular editing, trajectory memory, or hierarchical scaffold library from MolHop-4B. (a) Training reward. (b) Validation joint@H1. (c) Final joint@H1 on ChEMBL. (d) Final joint@H1 on Enamine.}
% \label{fig:4}
% \end{figure}

\subsubsection{OOD Optimization beyond Scaffold Library Coverage}

\begin{figure}[!htbp]
\centering
\renewcommand{\thefigure}{5}
\includegraphics[width=\linewidth,height=0.70\textheight,keepaspectratio]{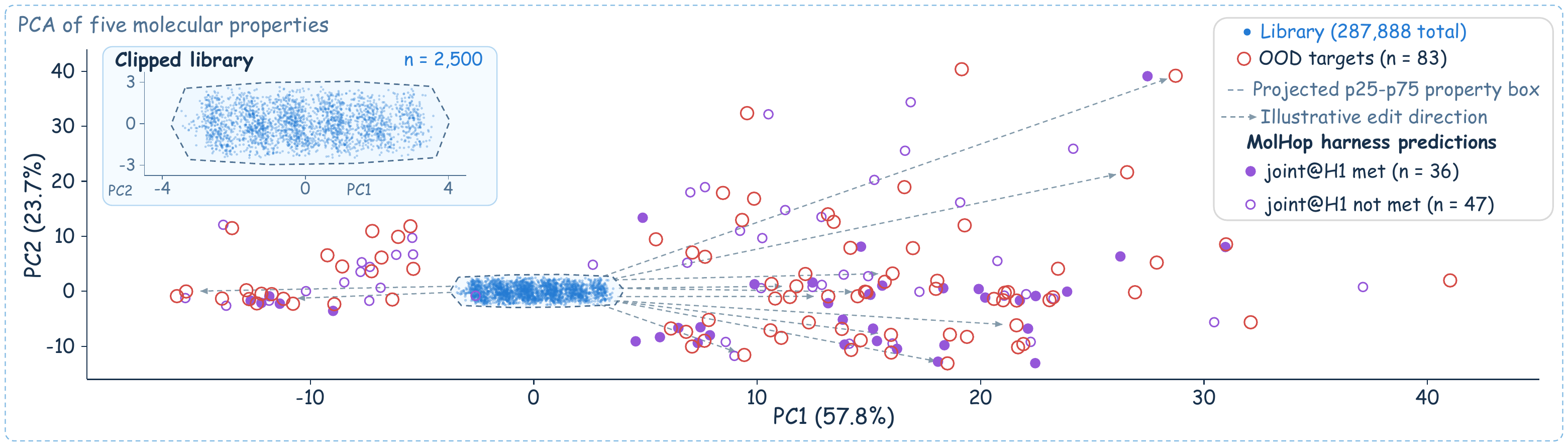}
\caption{OOD targets, restricted-library samples, and DeepSeek outputs in the property PCA space. PCA projection based on five molecular properties, logP, TPSA, MW, Labute ASA, and MolMR. Red open circles denote OOD targets, blue points denote library samples, and purple points denote MolHop harness predictions, with filled and open purple markers indicating joint@H1 met and joint@H1 not met outputs, respectively.}
\label{fig:D1}
\end{figure}

To further evaluate whether MolHop can optimize beyond the directly accessible scaffold space, we construct a restricted scaffold library by masking the central property region of the original library. As shown in Figure \ref{fig:D1}, the restricted library covers only a limited property region, while the OOD targets lie outside this visible space. Despite the lack of close scaffold retrieval candidates, DeepSeek + MolHop harness achieves 43.4\% joint@H1 on these targets, substantially improving over scaffold retrieval alone. This result indicates that the harness enables the agent to continue property optimization through molecular editing beyond the coverage of the visible scaffold library. Detailed construction of the restricted library and full results are provided in Appendix \ref{sec:app-ood}.

\section{Conclusion}

This work studies cross-scaffold molecular discovery as a search problem that requires both structural exploration and property optimization. Unlike conventional molecular optimization methods that mainly perform local modifications around existing structures, scaffold hopping requires agents to identify and navigate new structural regions while recovering target properties. To address this challenge, we propose MolHop, an agentic framework that organizes molecular scaffolds into a hierarchical search space and integrates scaffold navigation, molecular editing, and trajectory memory into a unified interaction environment.

Our experiments demonstrate that the MolHop harness substantially improves the cross-scaffold discovery ability of frozen LLM agents, showing that structured chemical search spaces and action interfaces can effectively support molecular exploration. Furthermore, agentic reinforcement learning enables models at different scales to learn search policies that coordinate scaffold selection and molecular editing under multi-objective constraints. These results highlight the importance of designing structured environments for molecular agents, where exploration decisions are explicitly represented rather than emerging only from local molecular transformations. We believe MolHop provides a step toward more capable molecular discovery agents that can navigate broader chemical spaces through learned interaction with structured chemical environments.

\FloatBarrier
\label{main-text-end}
\subsection*{AI use statement}

We used Cursor for coding assistance and Codex for language polishing and figure preparation. Grok-4.6 generated SFT demonstration trajectories from 2,000 starting points within the MolHop environment, which supplied molecular states and property feedback. The authors reviewed AI-assisted outputs and take responsibility for the final work.

\FloatBarrier
\bibliography{references}

\begin{thebibliography}{21}
\providecommand{\natexlab}[1]{#1}
\providecommand{\url}[1]{\texttt{#1}}
\expandafter\ifx\csname urlstyle\endcsname\relax
  \providecommand{\doi}[1]{doi: #1}\else
  \providecommand{\doi}{doi: \begingroup \urlstyle{rm}\Url}\fi

\bibitem[Abd et~al.(2026)Abd, Aggarwal, Algayres, Andreev, Bachem, Ballantyne, Brick, Carbune, Casbon, Chaturvedi, Chawla, Cotruta, Coucke, Culliton, Dadashi, Dixon, Elhawaty, Evci, Farabet, Ferret, Galgani, Girgin, Grill, Grootendorst, Guo, Hardin, He, Hernandez, Homburger, Hussenot, Ji, Joulin, Kamath, Kassraie, Lacombe, Lahoti, Liu, Martins, Martins, Matejovicova, Merhej, Momchev, Mondal, Mullins, Panyam, Pathak, Perrin, Pinto, Pot, Pouget, Ram{\'{e}}, Ramos, Reid, Rim, Rivi{\`{e}}re, Roth, Rouillard, Sanseviero, Sessa, Settle, Sinopalnikov, Smoot, Stanczyk, Steiner, Stewart, Tolstikhin, Tschannen, Tsitsulin, Vieillard, Wu, Xu, Yang, Yvinec, Zhang, Zhang, Zou, Aagnes, Abdelhamed, Ad{\'{a}}mek, Agrawal, Agrawal, Alabdulmohsin, Alayrac, Alon, Amarnath, Anand, Anastasiou, Ariafar, Aubet, Axiotis, Barbero, Barral, Bendebury, Bergmann, Bileschi, Black, Blondel, Borgeaud, and Brazinskas]{team2026gemma}
Sherif~El Abd, Vaibhav Aggarwal, Robin Algayres, Alek Andreev, Olivier Bachem, Ian Ballantyne, Cormac Brick, Victor Carbune, Michelle Casbon, Mayank Chaturvedi, Aditya Chawla, Victor Cotruta, Alice Coucke, Phil Culliton, Robert Dadashi, Lucas Dixon, Mohamed Elhawaty, Utku Evci, Cl{\'{e}}ment Farabet, Johan Ferret, Filippo Galgani, Sertan Girgin, Jean{-}Bastien Grill, Maarten Grootendorst, Jiaxian Guo, Cassidy Hardin, Yanzhang He, Steven~M. Hernandez, Omri Homburger, L{\'{e}}onard Hussenot, Ju{-}yeong Ji, Armand Joulin, Aishwarya Kamath, Parnian Kassraie, Olivier Lacombe, Preethi Lahoti, Ga{\"{e}}l Liu, Gus Martins, Luciano Martins, Tatiana Matejovicova, Ramona Merhej, Nikola Momchev, Sneha Mondal, Ryan Mullins, Sindhu~Raghuram Panyam, Shreya Pathak, Sarah Perrin, Andr{\'{e}}~Susano Pinto, Etienne Pot, Ang{\'{e}}line Pouget, Alexandre Ram{\'{e}}, Sabela Ramos, Douglas Reid, David Rim, Morgane Rivi{\`{e}}re, Karsten Roth, Louis Rouillard, Omar Sanseviero, Pier~Giuseppe Sessa, Shane Settle, Danila Sinopalnikov,
  Sara Smoot, Piotr Stanczyk, Andreas Steiner, Lawrence Stewart, Ilya~O. Tolstikhin, Michael Tschannen, Anton Tsitsulin, Nino Vieillard, Renjie Wu, Pingmei Xu, Haichuan Yang, Edouard Yvinec, Biao Zhang, Li~Zhang, Joe Zou, Nicolas Aagnes, Abdelrahman Abdelhamed, Jakub Ad{\'{a}}mek, Shivani Agrawal, Shubham Agrawal, Ibrahim Alabdulmohsin, Jean{-}Baptiste Alayrac, Uri Alon, Chandramouli Amarnath, Ankesh Anand, Chrysovalantis Anastasiou, Setareh Ariafar, Fran{\c{c}}ois{-}Xavier Aubet, Kyriakos Axiotis, Federico Barbero, Joelle~K. Barral, Alexei Bendebury, Urs Bergmann, Stanley Bileschi, Kat Black, Mathieu Blondel, Sebastian Borgeaud, and Arthur Brazinskas.
\newblock Gemma 4 technical report.
\newblock \emph{CoRR}, abs/2607.02770, 2026.
\newblock \doi{10.48550/ARXIV.2607.02770}.
\newblock URL \url{https://doi.org/10.48550/arXiv.2607.02770}.

\bibitem[Abdin et~al.(2024)Abdin, Aneja, Behl, Bubeck, Eldan, Gunasekar, Harrison, Hewett, Javaheripi, Kauffmann, Lee, Lee, Li, Liu, Mendes, Nguyen, Price, de~Rosa, Saarikivi, Salim, Shah, Wang, Ward, Wu, Yu, Zhang, and Zhang]{abdin2024phi}
Marah~I Abdin, Jyoti Aneja, Harkirat~S. Behl, S{\'{e}}bastien Bubeck, Ronen Eldan, Suriya Gunasekar, Michael Harrison, Russell~J. Hewett, Mojan Javaheripi, Piero Kauffmann, James~R. Lee, Yin~Tat Lee, Yuanzhi Li, Weishung Liu, Caio C.~T. Mendes, Anh Nguyen, Eric Price, Gustavo de~Rosa, Olli Saarikivi, Adil Salim, Shital Shah, Xin Wang, Rachel Ward, Yue Wu, Dingli Yu, Cyril Zhang, and Yi~Zhang.
\newblock Phi-4 technical report.
\newblock \emph{CoRR}, abs/2412.08905, 2024.
\newblock \doi{10.48550/ARXIV.2412.08905}.
\newblock URL \url{https://doi.org/10.48550/arXiv.2412.08905}.

\bibitem[Brown et~al.(2019)Brown, Fiscato, Segler, and Vaucher]{brown2019guacamol}
Nathan Brown, Marco Fiscato, Marwin~H.S. Segler, and Alain~C. Vaucher.
\newblock Guacamol: Benchmarking models for de novo molecular design.
\newblock \emph{Journal of Chemical Information and Modeling}, 59\penalty0 (3):\penalty0 1096--1108, 03 2019.
\newblock ISSN 1549-9596.
\newblock \doi{10.1021/acs.jcim.8b00839}.
\newblock URL \url{https://doi.org/10.1021/acs.jcim.8b00839}.

\bibitem[Ertl \& Schuffenhauer(2009)Ertl and Schuffenhauer]{ertl2009estimation}
Peter Ertl and Ansgar Schuffenhauer.
\newblock Estimation of synthetic accessibility score of drug-like molecules based on molecular complexity and fragment contributions.
\newblock \emph{J. Cheminformatics}, 1:\penalty0 8, 2009.
\newblock \doi{10.1186/1758-2946-1-8}.
\newblock URL \url{https://doi.org/10.1186/1758-2946-1-8}.

\bibitem[Gao et~al.(2022)Gao, Fu, Sun, and Coley]{gao2022sample}
Wenhao Gao, Tianfan Fu, Jimeng Sun, and Connor Coley.
\newblock Sample efficiency matters: A benchmark for practical molecular optimization.
\newblock In S.~Koyejo, S.~Mohamed, A.~Agarwal, D.~Belgrave, K.~Cho, and A.~Oh (eds.), \emph{Advances in Neural Information Processing Systems}, volume~35, pp.\  21342--21357. Curran Associates, Inc., 2022.
\newblock \doi{10.52202/068431-1551}.
\newblock URL \url{https://proceedings.neurips.cc/paper_files/paper/2022/file/8644353f7d307baaf29bc1e56fe8e0ec-Paper-Datasets_and_Benchmarks.pdf}.

\bibitem[Gaulton et~al.(2012)Gaulton, Bellis, Bento, Chambers, Davies, Hersey, Light, McGlinchey, Michalovich, Al-Lazikani, and Overington]{gaulton2012chembl}
Anna Gaulton, Louisa~J. Bellis, A.~Patricia Bento, Jon Chambers, Mark Davies, Anne Hersey, Yvonne Light, Shaun McGlinchey, David Michalovich, Bissan Al-Lazikani, and John~P. Overington.
\newblock Chembl: a large-scale bioactivity database for drug discovery.
\newblock \emph{Nucleic Acids Research}, 40\penalty0 (D1):\penalty0 D1100--D1107, 01 2012.
\newblock ISSN 0305-1048.
\newblock \doi{10.1093/nar/gkr777}.
\newblock URL \url{https://doi.org/10.1093/nar/gkr777}.

\bibitem[Guo et~al.(2025)Guo, Yang, Zhang, Song, Wang, Zhu, Xu, Zhang, Ma, Bi, Zhang, Yu, Wu, Wu, Gou, Shao, Li, Gao, Liu, Xue, Wang, Wu, Feng, Lu, Zhao, Deng, Ruan, Dai, Chen, Ji, Li, Lin, Dai, Luo, Hao, Chen, Li, Zhang, Xu, Ding, Gao, Qu, Li, Guo, Li, Chen, Yuan, Tu, Qiu, Li, Cai, Ni, Liang, Chen, Dong, Hu, You, Gao, Guan, Huang, Yu, Wang, Zhang, Zhao, Wang, Zhang, Xu, Xia, Zhang, Zhang, Tang, Zhou, Li, Wang, Li, Tian, Huang, Zhang, Wang, Chen, Du, Ge, Zhang, Pan, Wang, Chen, Jin, Chen, Lu, Zhou, Chen, Ye, Wang, Yu, Zhou, Pan, Li, Zhou, Wu, Yun, Pei, Sun, Wang, Zeng, Liu, Liang, Gao, Yu, Zhang, Xiao, An, Liu, Wang, Chen, Nie, Cheng, Liu, Xie, Liu, Yang, Li, Su, Lin, Li, Jin, Shen, Chen, Sun, Wang, Song, Zhou, Wang, Shan, Li, Wang, Wei, Zhang, Xu, Li, Zhao, Sun, Wang, Yu, Zhang, Shi, Xiong, He, Piao, Wang, Tan, Ma, Liu, Guo, Ou, Wang, Gong, Zou, He, Xiong, Luo, You, Liu, Zhou, Zhu, Huang, Li, Zheng, Zhu, Ma, Tang, Zha, Yan, Ren, Ren, Sha, Fu, Xu, Xie, Zhang, Hao, Ma, Yan, Wu, Gu, Zhu, Liu, Li, Xie, Song,
  Pan, Huang, Xu, Zhang, and Zhang]{guo2025deepseek}
Daya Guo, Dejian Yang, Haowei Zhang, Junxiao Song, Peiyi Wang, Qihao Zhu, Runxin Xu, Ruoyu Zhang, Shirong Ma, Xiao Bi, Xiaokang Zhang, Xingkai Yu, Yu~Wu, Z.~F. Wu, Zhibin Gou, Zhihong Shao, Zhuoshu Li, Ziyi Gao, Aixin Liu, Bing Xue, Bingxuan Wang, Bochao Wu, Bei Feng, Chengda Lu, Chenggang Zhao, Chengqi Deng, Chong Ruan, Damai Dai, Deli Chen, Dongjie Ji, Erhang Li, Fangyun Lin, Fucong Dai, Fuli Luo, Guangbo Hao, Guanting Chen, Guowei Li, H.~Zhang, Hanwei Xu, Honghui Ding, Huazuo Gao, Hui Qu, Hui Li, Jianzhong Guo, Jiashi Li, Jingchang Chen, Jingyang Yuan, Jinhao Tu, Junjie Qiu, Junlong Li, J.~L. Cai, Jiaqi Ni, Jian Liang, Jin Chen, Kai Dong, Kai Hu, Kaichao You, Kaige Gao, Kang Guan, Kexin Huang, Kuai Yu, Lean Wang, Lecong Zhang, Liang Zhao, Litong Wang, Liyue Zhang, Lei Xu, Leyi Xia, Mingchuan Zhang, Minghua Zhang, Minghui Tang, Mingxu Zhou, Meng Li, Miaojun Wang, Mingming Li, Ning Tian, Panpan Huang, Peng Zhang, Qiancheng Wang, Qinyu Chen, Qiushi Du, Ruiqi Ge, Ruisong Zhang, Ruizhe Pan, Runji Wang, R.~J.
  Chen, R.~L. Jin, Ruyi Chen, Shanghao Lu, Shangyan Zhou, Shanhuang Chen, Shengfeng Ye, Shiyu Wang, Shuiping Yu, Shunfeng Zhou, Shuting Pan, S.~S. Li, Shuang Zhou, Shaoqing Wu, Tao Yun, Tian Pei, Tianyu Sun, T.~Wang, Wangding Zeng, Wen Liu, Wenfeng Liang, Wenjun Gao, Wenqin Yu, Wentao Zhang, W.~L. Xiao, Wei An, Xiaodong Liu, Xiaohan Wang, Xiaokang Chen, Xiaotao Nie, Xin Cheng, Xin Liu, Xin Xie, Xingchao Liu, Xinyu Yang, Xinyuan Li, Xuecheng Su, Xuheng Lin, X.~Q. Li, Xiangyue Jin, Xiaojin Shen, Xiaosha Chen, Xiaowen Sun, Xiaoxiang Wang, Xinnan Song, Xinyi Zhou, Xianzu Wang, Xinxia Shan, Y.~K. Li, Y.~Q. Wang, Y.~X. Wei, Yang Zhang, Yanhong Xu, Yao Li, Yao Zhao, Yaofeng Sun, Yaohui Wang, Yi~Yu, Yichao Zhang, Yifan Shi, Yiliang Xiong, Ying He, Yishi Piao, Yisong Wang, Yixuan Tan, Yiyang Ma, Yiyuan Liu, Yongqiang Guo, Yuan Ou, Yuduan Wang, Yue Gong, Yuheng Zou, Yujia He, Yunfan Xiong, Yuxiang Luo, Yuxiang You, Yuxuan Liu, Yuyang Zhou, Y.~X. Zhu, Yanping Huang, Yaohui Li, Yi~Zheng, Yuchen Zhu, Yunxian Ma, Ying
  Tang, Yukun Zha, Yuting Yan, Z.~Z. Ren, Zehui Ren, Zhangli Sha, Zhe Fu, Zhean Xu, Zhenda Xie, Zhengyan Zhang, Zhewen Hao, Zhicheng Ma, Zhigang Yan, Zhiyu Wu, Zihui Gu, Zijia Zhu, Zijun Liu, Zilin Li, Ziwei Xie, Ziyang Song, Zizheng Pan, Zhen Huang, Zhipeng Xu, Zhongyu Zhang, and Zhen Zhang.
\newblock Deepseek-r1 incentivizes reasoning in llms through reinforcement learning.
\newblock \emph{Nature}, 645\penalty0 (8081):\penalty0 633–638, 2025.
\newblock ISSN 1476-4687.
\newblock \doi{10.1038/s41586-025-09422-z}.
\newblock URL \url{http://dx.doi.org/10.1038/s41586-025-09422-z}.

\bibitem[Jensen(2019)]{jensen2019graph}
Jan~H. Jensen.
\newblock A graph-based genetic algorithm and generative model/monte carlo tree search for the exploration of chemical space.
\newblock \emph{Chemical Science}, 10\penalty0 (12):\penalty0 3567--3572, 03 2019.
\newblock ISSN 2041-6520.
\newblock \doi{10.1039/c8sc05372c}.
\newblock URL \url{https://doi.org/10.1039/c8sc05372c}.

\bibitem[Kim et~al.(2024)Kim, Kim, Choi, and Park]{kim2024genetic}
Hyeonah Kim, Minsu Kim, Sanghyeok Choi, and Jinkyoo Park.
\newblock Genetic-guided gflownets for sample efficient molecular optimization.
\newblock In A.~Globerson, L.~Mackey, D.~Belgrave, A.~Fan, U.~Paquet, J.~Tomczak, and C.~Zhang (eds.), \emph{Advances in Neural Information Processing Systems}, volume~37, pp.\  42618--42648. Curran Associates, Inc., 2024.
\newblock \doi{10.52202/079017-1350}.
\newblock URL \url{https://proceedings.neurips.cc/paper_files/paper/2024/file/4b25c000967af9036fb9b207b198a626-Paper-Conference.pdf}.

\bibitem[Lee et~al.(2025)Lee, Kreis, Veccham, Liu, Reidenbach, Peng, Paliwal, Nie, and Vahdat]{lee2025genmol}
Seul Lee, Karsten Kreis, Srimukh~Prasad Veccham, Meng Liu, Danny Reidenbach, Yuxing Peng, Saee~Gopal Paliwal, Weili Nie, and Arash Vahdat.
\newblock {G}en{M}ol: A drug discovery generalist with discrete diffusion.
\newblock In Aarti Singh, Maryam Fazel, Daniel Hsu, Simon Lacoste-Julien, Felix Berkenkamp, Tegan Maharaj, Kiri Wagstaff, and Jerry Zhu (eds.), \emph{Proceedings of the 42nd International Conference on Machine Learning}, volume 267 of \emph{Proceedings of Machine Learning Research}, pp.\  33205--33226. PMLR, 13--19 Jul 2025.
\newblock URL \url{https://proceedings.mlr.press/v267/lee25o.html}.

\bibitem[Li et~al.(2026{\natexlab{a}})Li, Cao, Peng, Liu, Feng, Wang, Yan, Tian, Li, and Yuan]{li2026agentic}
Hao Li, He~Cao, Shenyao Peng, Zijing Liu, Bin Feng, Yu~Wang, Zhiyuan Yan, Yonghong Tian, Yu~Li, and Li~Yuan.
\newblock Agentic reinforcement learning empowers next-generation chemical language models for molecular design and synthesis.
\newblock \emph{CoRR}, abs/2601.17687, 2026{\natexlab{a}}.
\newblock \doi{10.48550/ARXIV.2601.17687}.
\newblock URL \url{https://doi.org/10.48550/arXiv.2601.17687}.

\bibitem[Li et~al.(2026{\natexlab{b}})Li, Hou, Vinh, Raj, Guo, and Yang]{li2026reinforcement}
Tao Li, Kaiyuan Hou, Tuan Vinh, Monika Raj, Zhichun Guo, and Carl Yang.
\newblock Reinforcement learning with llm-guided action spaces for synthesizable lead optimization.
\newblock \emph{CoRR}, abs/2604.07669, 2026{\natexlab{b}}.
\newblock \doi{10.48550/ARXIV.2604.07669}.
\newblock URL \url{https://doi.org/10.48550/arXiv.2604.07669}.

\bibitem[Olivecrona et~al.(2017)Olivecrona, Blaschke, Engkvist, and Chen]{olivecrona2017molecular}
Marcus Olivecrona, Thomas Blaschke, Ola Engkvist, and Hongming Chen.
\newblock Molecular de-novo design through deep reinforcement learning.
\newblock \emph{J. Cheminformatics}, 9\penalty0 (1):\penalty0 48:1--48:14, 2017.
\newblock \doi{10.1186/S13321-017-0235-X}.
\newblock URL \url{https://doi.org/10.1186/s13321-017-0235-x}.

\bibitem[Team(2024)]{grattafiori2024llama}
Llama Team.
\newblock The llama 3 herd of models.
\newblock \emph{CoRR}, abs/2407.21783, 2024.
\newblock \doi{10.48550/ARXIV.2407.21783}.
\newblock URL \url{https://doi.org/10.48550/arXiv.2407.21783}.

\bibitem[Team(2026)]{qwen2026qwen35}
Qwen Team.
\newblock Qwen3.5: Towards native multimodal agents, February 2026.
\newblock URL \url{https://qwen.ai/blog?id=qwen3.5}.

\bibitem[Xu et~al.(2026)Xu, Li, Lin, Xue, Xian, Xu, Wu, Zhang, Deng, Yu, Jin, Lin, Dong, Wang, Feng, Lu, Zhao, Deng, Zhang, Xu, Zhao, Shao, Wang, Zhang, Dai, Yang, Chen, Huang, Wu, Li, Li, Fu, Zhou, Zhou, Lin, Yuan, Xia, Dai, Hao, Li, Chen, Cao, Fan, Meng, Li, Zhang, Ma, Shen, Li, Yu, Zhang, Deng, Xu, Xu, Zhong, Guo, Jiang, Li, Qin, Wen, Liang, Huang, Liu, Zhang, Luo, Yang, Xu, Yuan, Huang, Luo, Cai, Chen, Ji, Zhang, Wang, Wu, Ding, Tang, Wang, Cao, Gao, Qu, Zeng, Yang, Jin, Zhang, Zou, Yu, Zhou, Chen, Huang, Zhao, Tang, Zhou, Tong, Li, Zhu, Wang, Ye, Li, Xu, Ding, Lu, Hu, Yan, Zhai, Chen, Hu, Zhou, Xu, Xiang, Yun, Yuan, Cheng, Zhu, Wang, Chen, Hu, Yu, Guo, Pei, Sun, Jiang, Qiu, Zhou, Liu, Li, Li, Song, Guo, Dong, Chen, Gao, Guan, Yuan, Hong, Xu, Zhao, Ji, Zhang, Zhou, Yu, Zhang, Wang, Zhang, Wang, Gao, Zhao, Xu, Guo, Luo, Fu, Deng, Wang, Zhang, Chen, Chen, Huang, Ma, Wang, Di, Mei, Ye, Cui, Zhang, Zhang, Tang, Zhang, Wei, Chen, Liu, Zhou, Xu, Yang, Wang, Chen, Shentu, Wang, Ning, Huang, Cong, Wang, Xin, Ren,
  Yan, Zhang, Kang, Tang, Wang, Li, Zhu, Li, Chen, Du, Guo, Xu, Ding, Hu, Tian, Yu, Zhu, Xu, Yang, Xia, Lu, Geng, Hong, Ge, Zhang, Sun, Pan, Wang, Chen, Xu, Tian, Shen, Zhang, X., Liu, Lu, Zhou, Chen, Cai, Nie, Chen, Hu, Lin, Ran, Liu, Jia, Bai, Feng, Xu, Liu, Hu, Ma, Wang, Feng, Gong, Lin, Yu, Zhou, Yang, Wang, Guo, Pan, Yu, Cao, Lin, Chen, Chen, Zhou, Ni, Yun, Jin, Pei, Ye, Lin, Ji, Cui, Yue, Yu, Xiong, Zeng, Liu, Zhang, Xu, Zeng, Zhao, Liu, Liang, Pang, Luo, Yao, Gao, Shao, Yang, Zhang, Wang, Huang, Yan, Zhang, Gao, He, Li, Li, Wang, Zhang, Wei, Bi, Liu, Wang, Qu, Chen, Zhang, Nie, Zou, Li, Guo, Chu, Cheng, Liu, Xie, Xu, Liu, Liu, Yu, Li, Yao, Chen, Jiang, Yang, Yang, Chen, Wang, Zhong, Su, Liu, Lin, Fan, Zhao, Fu, Yan, Jiang, Wu, M., Wang, Gao, Yang, Zhang, Ma, Huang, Li, Li, Meng, Zhao, Sun, Wang, Ye, Yin, Wu, Qian, Tao, Yu, Zhang, Jiang, Wang, Ding, Shi, Peng, Zhai, Wu, Xiong, Wang, He, Zhou, Luo, Zhong, Wang, Wang, Zhang, Chen, Tan, Wei, Ma, Yang, Liu, Cai, Wei, Wang, Yang, Zhuo, Guo, Wu, Wu, Zhang,
  Bian, Cheng, Ou, Sun, Xu, Sun, Li, Liu, Yao, Han, Wang, Wu, Meng, Zou, Li, Wang, Xiao, Xiong, Chen, Cao, Wang, Chen, Zhang, Lin, Xiao, Gu, Chen, Huang, Luo, You, Chen, Xiang, Liu, Zhou, Zhou, Guo, Huang, Bai, Z., Ni, Wang, Zhao, Ren, Zhao, Sha, Wang, Zhang, Du, Fu, Xu, Xie, Liu, Zhang, Dong, Hao, Wang, Gou, Ma, Li, Shao, Huang, Li, Lu, Huang, Chen, Chen, Pan, Wu, Ren, He, Li, Zhang, Xu, Wang, Gu, Zhu, Zhang, Li, Hou, Lyu, Wang, Xie, Zhang, Gao, Pan, Li, Yao, Chen, Wu, Ling, Hou, Chen, Li, Qi, Ji, Wei, Xia, Xie, Tan, Guo, Zhai, Zhou, Tan, Li, Luo, Song, Cai, Liang, Zhou, Gao, Shao, Chen, Yang, Chen, Zhang, Zhou, Wang, Liu, Sun, Ling, Zheng, Yang, Xu, Su, Xia, Ding, Zhuo, Ma, Zhu, Cai, Yao, Zhang, Li, Lin, Wang, Zhang, Li, Wang, Yin, Han, Cao, Wang, Ma, Wang, Lin, Sun, Zhang, Ying, Xiang, Wang, Mao, Jiang, Jin, Chen, Tao, Sun, Wu, Zou, Lei, Zhang, Sun, Yin, Xiao, An, Li, Wang, Lin, Hou, Lin, Meng, Huang, Peng, Li, Zhang, Sun, Wang, Ye, Zhang, Zhang, Cao, Chen, Zhou, Xu, Xia, Xu, Shao, Zhang, Ma, Tang, Lou,
  Chen, Piao, Chen, Xiong, Xuan, Yang, Xu, Zha, Ma, Lin, Yan, Xie, Sheng, Zhu, Zhang, Ju, Lin, Gao, Sun, Yan, Wu, Wang, Qu, Yan, and Wan]{xu2026deepseek}
DeepSeek-AI:~Anyi Xu, B.~Li, Bangcai Lin, Bing Xue, BingCheng Xian, Bingzheng Xu, Bochao Wu, Bowei Zhang, Boyi Deng, C.~C. Yu, Chao Jin, Chaofan Lin, Chen Dong, Chenbing Wang, Chenfan Feng, Chengda Lu, Chenggang Zhao, Chengqi Deng, Chengyuan Zhang, Chenhao Xu, Chenqi Zhao, Chenze Shao, Chuhao Wang, Chuqi Zhang, Damai Dai, Dejian Yang, Deli Chen, Di~Huang, Di~Wu, Donghao Li, Erhang Li, Eric Fu, F.~Zhou, Fangwei Zhou, Fangyun Lin, Fangzhou Yuan, Feiyu Xia, Fucong Dai, Guangbo Hao, Guanglin Li, Guanting Chen, Guoai Cao, Guofan Fan, Guolai Meng, Guowei Li, Haichuan Zhang, Haiyang Ma, Haiyang Shen, Han Li, Han Yu, Han Zhang, Hangyuan Deng, Hanwei Xu, Hanxiang Xu, Hanxun Zhong, Hao Guo, Hao Jiang, Hao Li, Hao Qin, Haodong Wen, Haofen Liang, Haofeng Huang, Haohua Liu, Haoling Zhang, Haoming Luo, Haoran Yang, Haotian Xu, Haotian Yuan, Haoting Huang, Haowen Luo, Haoyang Cai, Haoyu Chen, Haozhe Ji, Hengran Zhang, Hengrui Wang, Hengxu Wu, Honghui Ding, Hongxuan Tang, Huadong Wang, Huanqi Cao, Huazuo Gao, Hui Qu, Hui
  Zeng, J.~Yang, J.~H. Jin, J.~H. Zhang, J.~X. Zou, Jia Yu, Jiahui Zhou, Jiajun Chen, Jialiang Huang, Jialin Zhao, Jiamin Tang, Jian Zhou, Jianan Tong, Jianwen Li, Jiaqi Zhu, Jiarui Wang, Jiasheng Ye, Jiashi Li, Jiaxin Xu, Jiaying Ding, Jibai Lu, Jiewen Hu, Jin Yan, Jincheng Zhai, Jingchang Chen, Jingcheng Hu, Jingli Zhou, Jingsheng Xu, Jingting Xiang, Jingyan Yun, Jingyang Yuan, Jingyuan Cheng, Jinhua Zhu, Jinpeng Wang, Jinyi Chen, Jinyi Hu, Jiping Yu, Jueliang Guo, Junbo Pei, Junbo Sun, Junguang Jiang, Junjie Qiu, Junkang Zhou, Junqi Liu, Junren Li, Junxian Li, Junxiao Song, Junyi Guo, Kai Dong, Kaifeng Chen, Kaige Gao, Kang Guan, Kangdong Yuan, Ke~Hong, Ke~Xu, Kefan Zhao, Kexin Ji, Kexin Zhang, Kexing Zhou, Kuai Yu, Lan Zhang, Lean Wang, Lecong Zhang, Lei Wang, Letian Gao, Liang Zhao, Liansheng Xu, Lihua Guo, Lingxiao Luo, Lingyue Fu, Litao Deng, Litong Wang, Liyue Zhang, Longhao Chen, Lu~Chen, Luotian Huang, Luyao Ma, Luyao Wang, M.~S. Di, Max Mei, Menghao Ye, Miao Cui, Mingchuan Zhang, Minghua Zhang,
  Minghui Tang, Mingjing Zhang, Mingqi Wei, Mingshu Chen, Mingxing Liu, Mingxu Zhou, Mingyu Xu, Mingyu Yang, Mingze Wang, Muyang Chen, Ni~Shentu, Ning Wang, Niufang Ning, Panpan Huang, Peixin Cong, Peiyi Wang, Peiyuan Xin, Pengfei Ren, Pengfei Yan, Pengle Zhang, Qi~Kang, Qi~Tang, Qiancheng Wang, Qiang Li, Qihao Zhu, Qingyang Li, Qinyu Chen, Qiushi Du, Qizhou Guo, Rongxian Xu, Rui Ding, Rui Hu, Rui Tian, Rui Yu, Ruidong Zhu, Ruifan Xu, Ruihan Yang, Ruihang Xia, Ruijie Lu, Ruilin Geng, Ruipeng Hong, Ruiqi Ge, Ruisong Zhang, Ruize Sun, Ruizhe Pan, Runji Wang, Runqian Chen, Runxin Xu, Ruohong Tian, Ruomeng Shen, Ruoyu Zhang, Ryan X., S.~H. Liu, Shanghao Lu, Shangyan Zhou, Shanhuang Chen, Shaofei Cai, Shaoheng Nie, Shaoyuan Chen, Shengding Hu, Shengkai Lin, Shengwen Ran, Shengyu Liu, Shengyuan Jia, Shi Bai, Shi Feng, Shicheng Xu, Shichun Liu, Shiqiang Hu, Shirong Ma, Shiyu Wang, Shiyuan Feng, Shufan Gong, Shuhan Lin, Shuiping Yu, Shunfeng Zhou, Shuo Yang, Shuomeng Wang, Shuting Guo, Shuting Pan, Shuying Yu, Sinuo
  Cao, Siyi Lin, Sizhe Chen, Songyang Chen, Songyang Zhou, Tao Ni, Tao Yun, Tian Jin, Tian Pei, Tian Ye, Tianle Lin, Tianran Ji, Tianyi Cui, Tianyuan Yue, Tingting Yu, Tongrui Xiong, Wangding Zeng, Wei Liu, Wei Zhang, Weibin Xu, Weihao Zeng, Weilin Zhao, Wen Liu, Wenfeng Liang, Wenjie Pang, Wenjing Luo, Wenjing Yao, Wenjun Gao, Wenkai Shao, Wenkai Yang, Wenli Zhang, Wenlu Wang, Wenlve Huang, Wenqian Yan, Wentao Zhang, Xi~Gao, Xiang He, Xiang Li, Xiangli Li, Xiangwen Wang, Xiangying Zhang, Xiankui Wei, Xiao Bi, Xiaodong Liu, Xiaohan Wang, Xiaojian Qu, Xiaokang Chen, Xiaokang Zhang, Xiaotao Nie, Xiaoyao Zou, Xiaoyuan Li, Xicheng Guo, Xieting Chu, Xin Cheng, Xin Liu, Xin Xie, Xinbo Xu, Xingchao Liu, Xingchen Liu, Xingkai Yu, Xingyou Li, Xintong Yao, Xinyang Chen, Xinyong Jiang, Xinyu Yang, Xinyu Yang, Xu~Chen, Xuanyu Wang, Xubei Zhong, Xuecheng Su, Xuejie Liu, Xuheng Lin, Xujie Fan, Xuncheng Zhao, Xuwei Fu, Y.~C. Yan, Y.~H. Jiang, Y.~T. Wu, Y.~W. M., Y.~Z. Wang, Yafei Gao, Yang Yang, Yang Zhang, Yanru Ma, Yanwen
  Huang, Yao Li, Yao Li, Yao Meng, Yao Zhao, Yaofeng Sun, Yaohui Wang, Yaoyang Ye, Yehang Yin, Yexinrui Wu, Yi~Qian, Yi~Tao, Yi~Yu, Yichao Zhang, Yichen Jiang, Yicheng Wang, Yifan Ding, Yifan Shi, Yifeng Peng, Yifeng Zhai, Yijia Wu, Yiliang Xiong, Yilun Wang, Ying He, Ying Zhou, Yingjia Luo, Yinmin Zhong, Yiping Wang, Yisong Wang, Yixiang Zhang, Yixiao Chen, Yixuan Tan, Yixuan Wei, Yiyang Ma, Yiyao Yang, Yiyuan Liu, Yizai Cai, Yizhen Wei, Yizhi Wang, Yonglun Yang, Yongqi Zhuo, Yongqiang Guo, Yongtong Wu, Yu~Wu, Yu~Zhang, Yuan Bian, Yuan Cheng, Yuan Ou, Yuan Sun, Yuanfan Xu, Yuanhang Sun, Yuanhao Li, Yuchen Liu, Yuchen Yao, Yudong Han, Yuduan Wang, Yuhan Wu, Yuhao Meng, Yuheng Zou, YuKun Li, Yunchuan Wang, Yunfan Xiao, Yunfan Xiong, Yupeng Chen, Yuqian Cao, Yuqian Wang, Yuqing Chen, Yushun Zhang, Yutong Lin, Yuwei Xiao, Yuxian Gu, Yuxiang Chen, Yuxiang Huang, Yuxiang Luo, Yuxiang You, Yuxin Chen, Yuxin Xiang, Yuxuan Liu, Yuxuan Zhou, Yuyang Zhou, Yuzhe Guo, Yuzhen Huang, Yuzhuo Bai, Z.~Y. Z., Zanlin Ni, Zehao
  Wang, Zehua Zhao, Zehui Ren, Zejun Zhao, Zhangli Sha, Zhanying Wang, Zhaochen Zhang, Zhaoshuai Du, Zhe Fu, Zhean Xu, Zhenda Xie, Zheng Liu, Zhengyan Zhang, Zhenhua Dong, Zhewen Hao, Zhibang Wang, Zhibin Gou, Zhicheng Ma, Zhihao Li, Zhihong Shao, Zhihuan Huang, Zhijie Li, Zhirui Lu, Zhixian Huang, Zhixuan Chen, Zhixuan Chen, Zhixuan Pan, Zhiyu Wu, Zhizhou Ren, Zhu He, Zhuoshu Li, Zhuping Zhang, Zian Xu, Zihao Wang, Zihui Gu, Zijia Zhu, Zili Zhang, Zilin Li, Zilong Hou, Zilong Lyu, Ziqiao Wang, Ziwei Xie, Ziya Zhang, Ziyi Gao, Zizheng Pan, Zonglin Li, Zongqing Yao, Zui Chen, Zuofan Wu, Chenchen Ling, Chengyu Hou, Chong Chen, D.~Li, Di~Qi, Dongjie Ji, Fang Wei, Fanyi Xia, Fei Xie, Feiyi Tan, Hailong Guo, Haiyan Zhai, Hui Zhou, Huihui Tan, Huijie Li, Jia Luo, Jia Song, Jialu Cai, Jian Liang, Jiangting Zhou, Jiaqi Gao, Jiayi Shao, Jie Chen, Jieyu Yang, Jin Chen, Jingde Zhang, Jingzi Zhou, Jinqian Wang, Jinyang Liu, JinZhao Sun, Junhua Ling, Junmin Zheng, Kaicheng Yang, Ke~Xu, Le~Su, Leyi Xia, Liangfeng Ding, Lin
  Zhuo, Linwang Ma, Linyan Zhu, Liyu Cai, Luqi Yao, M.~K. Zhang, Meng Li, Miao Lin, Miaojun Wang, Min Zhang, Mingming Li, Mingming Wang, Mingze Yin, Minmin Han, Nan Cao, Ning Wang, Ningxin Ma, Panpan Wang, Peihan Lin, Peng Sun, Peng Zhang, Qian Ying, Qiang Xiang, Qiao Wang, Qingmiao Mao, Qiwei Jiang, Rongli Jin, Ruyi Chen, Sha Tao, Shangmian Sun, Shaoqing Wu, Shichao Zou, Si~Lei, Tianyang Zhang, Tianyu Sun, Tingting Yin, W.~L. Xiao, Wei An, Wei Li, Wei Wang, Weiwei Lin, Wenqing Hou, X.~Lin, Xiangfei Meng, Xianzhu Huang, Xiao Peng, Xiaoqian Li, Xiaoting Zhang, Xiaowen Sun, Xiaoxiang Wang, Xiaoyu Ye, Xinrou Zhang, Xinyu Zhang, Xue Cao, Xueyin Chen, Yanan Zhou, Yanhong Xu, Yao Xia, Yao Xu, Yi~Shao, Yihong Zhang, Yiling Ma, Ying Tang, Yining Lou, Yiru Chen, Yishi Piao, Yixuan Chen, Yong Xiong, Yuchen Xuan, Yuehan Yang, Yuer Xu, Yukun Zha, Yunxian Ma, Yuping Lin, Yuting Yan, Yutong Xie, Yuwen Sheng, Yuxuan Zhu, Zekai Zhang, Zhe Ju, Zhenzhen Lin, Zheren Gao, Zheyang Sun, Zhigang Yan, Zhongyu Wu, Zi~Wang, Zihua Qu,
  Ziling Yan, and Ziyi Wan.
\newblock Deepseek-v4.1-flash: Pushing the limits of kv cache compression, 2026.
\newblock URL \url{https://arxiv.org/abs/2609.19969}.

\bibitem[Yang et~al.(2025)Yang, Chen, Xie, Gao, Shen, Liu, Yang, Wang, Fu, and Li]{yang2025molact}
Zhuo Yang, Yeyun Chen, Jiaqing Xie, Ben Gao, Shuaike Shen, Wanhao Liu, Liujia Yang, Beilun Wang, Tianfan Fu, and Yuqiang Li.
\newblock Molact: An agentic {RL} framework for molecular editing and property optimization.
\newblock \emph{CoRR}, abs/2512.20135, 2025.
\newblock \doi{10.48550/ARXIV.2512.20135}.
\newblock URL \url{https://doi.org/10.48550/arXiv.2512.20135}.

\bibitem[Zhang et~al.(2024)Zhang, Liu, Tan, Chen, Yan, Yan, Li, Huang, Yue, Zhou, Zhang, Su, Zhong, Li, and Ouyang]{zhang2024chemllm}
Di~Zhang, Wei Liu, Qian Tan, Jingdan Chen, Hang Yan, Yuliang Yan, Jiatong Li, Weiran Huang, Xiangyu Yue, Dongzhan Zhou, Shufei Zhang, Mao Su, Hansen Zhong, Yuqiang Li, and Wanli Ouyang.
\newblock Chemllm: {A} chemical large language model.
\newblock \emph{CoRR}, abs/2402.06852, 2024.
\newblock \doi{10.48550/ARXIV.2402.06852}.
\newblock URL \url{https://doi.org/10.48550/arXiv.2402.06852}.

\bibitem[Zhao et~al.(2025)Zhao, Ma, Chen, Sun, Li, Xia, Chen, Xu, Zhu, Zhu, et~al.]{zhao2025developing}
Zihan Zhao, Da~Ma, Lu~Chen, Liangtai Sun, Zihao Li, Yi~Xia, Bo~Chen, Hongshen Xu, Zichen Zhu, Su~Zhu, et~al.
\newblock Developing chemdfm as a large language foundation model for chemistry.
\newblock \emph{Cell Reports Physical Science}, 6\penalty0 (4), 2025.
\newblock URL \url{https://doi.org/10.1016/j.xcrp.2025.102523}.

\bibitem[Zhao et~al.(2026)Zhao, Wan, Chen, Lin, Yu, Zhang, Ma, Zhu, Zhang, Wang, Dai, Wen, Chen, Chen, and Yu]{zhao2025chemdfm}
Zihan Zhao, Ziping Wan, Lu~Chen, Xuanze Lin, Shiyang Yu, Situo Zhang, Da~Ma, Zichen Zhu, Danyang Zhang, Huayang Wang, Zhongyang Dai, Liyang Wen, Bo~Chen, Xin Chen, and Kai Yu.
\newblock Chemdfm-r: A chemical reasoning llm enhanced with atomized chemical knowledge, 2026.
\newblock URL \url{https://arxiv.org/abs/2507.21990}.

\bibitem[Zhou et~al.(2026)Zhou, Vadgama, Varambally, Eckmann, Gilson, and Yu]{zhou2026toolmol}
Andrew~Y. Zhou, Sharvaree Vadgama, Sumanth Varambally, Peter Eckmann, Michael~K. Gilson, and Rose Yu.
\newblock Toolmol: Evolutionary agentic framework for multi-objective drug discovery.
\newblock \emph{CoRR}, abs/2605.12784, 2026.
\newblock \doi{10.48550/ARXIV.2605.12784}.
\newblock URL \url{https://doi.org/10.48550/arXiv.2605.12784}.

\end{thebibliography}
\bibliographystyle{iclr2027_conference}

\clearpage
\appendix
\section*{Appendix}

\FloatBarrier
\section{Detailed Design}

\FloatBarrier
\subsection{Hierarchical Scaffold Space}\label{sec:app-scaffold-space}

The scaffold space of MolHop is constructed from the Enamine Screening Collection, which contains 4,805,572 commercial molecular records. We first standardize the molecules using RDKit, including molecular parsing, SaltType filtering, and retention of the largest organic fragment in multicomponent molecules. We then extract Bemis--Murcko scaffolds, convert the results to canonical SMILES, and deduplicate them, yielding 1,811,153 unique scaffold records.

To ensure consistency between the scaffold space and the subsequent editing environment, we further filter the structural types. The search space retains single-component connected scaffolds and removes records containing bridgehead or spiro structures. There are 87,329 scaffolds containing only spiro structures, 55,729 containing only bridgehead structures, and 2,480 containing both types, resulting in a total of 145,538 excluded records. The final set contains 1,665,615 valid scaffold leaf nodes.

\begin{figure}[!htbp]
\centering
\renewcommand{\thefigure}{A1}
\includegraphics[width=\linewidth,height=0.70\textheight,keepaspectratio]{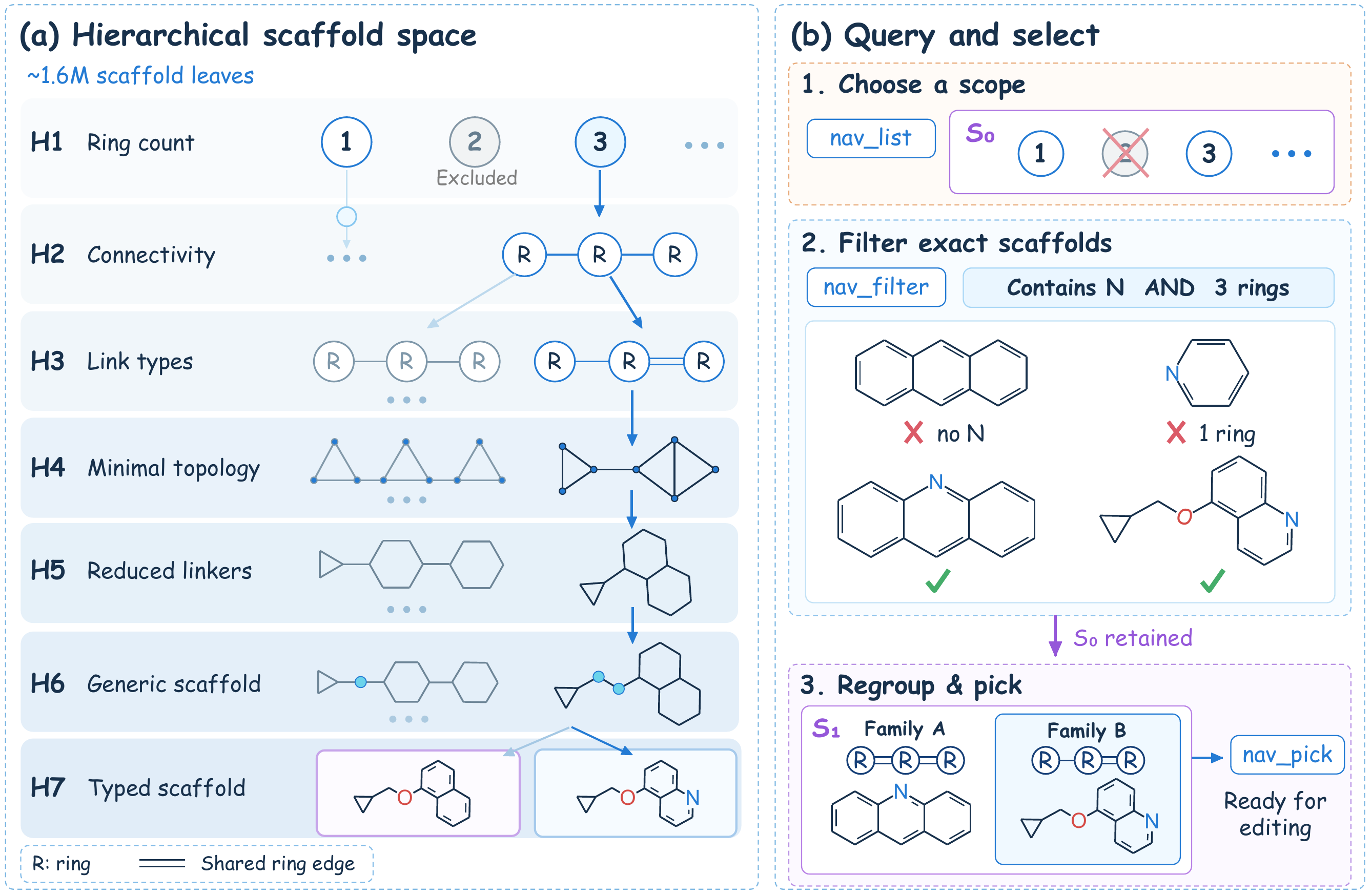}
\caption{Construction and query process of the hierarchical scaffold space. (a) Coarse-to-fine scaffold abstraction from H1 ring-level categories to H7 typed scaffolds. (b) Example scaffold query process, including scope selection, exact scaffold filtering, regrouping, and scaffold selection.}
\label{fig:A1}
\end{figure}

For each H7 exact scaffold, MolHop generates signatures for the H1-H7 hierarchical scaffold representation according to deterministic graph abstraction rules and establishes parent--child hierarchical relationships based on shared signatures. As shown in Figure \ref{fig:A1}(a), this hierarchy progressively restores connectivity, topological details, and chemical type information from coarse-grained ring structures, ultimately yielding a complete typed Murcko scaffold. The hierarchical representation is generated by predefined graph abstraction rules and does not rely on clustering with continuous similarity thresholds. The definitions and node counts at each level are shown in Table \ref{tab:A1}.

After the hierarchy is constructed, each H7 scaffold serves as a selectable leaf node. During search, the agent performs scaffold queries and scaffold selection based on hierarchical nodes and loads the selected H7 scaffold into the editing environment, as shown in Figure \ref{fig:A1}(b). For cross-scaffold tasks, MolHop records the H1 region of the reference molecule and checks the region of the final candidate at the end of the episode.

\begingroup
\renewcommand{\thetable}{A1}
\begin{longtable}{@{}>{\raggedright\arraybackslash}p{\dimexpr 0.100\linewidth-2\tabcolsep\relax}>{\raggedright\arraybackslash}p{\dimexpr 0.280\linewidth-2\tabcolsep\relax}>{\raggedright\arraybackslash}p{\dimexpr 0.460\linewidth-2\tabcolsep\relax}>{\raggedright\arraybackslash}p{\dimexpr 0.160\linewidth-2\tabcolsep\relax}@{}}
\caption{Hierarchical Scaffold Space Statistics}\label{tab:A1}\\
\toprule
Level & Retained structural information & Abstraction rule & Number of nodes \\

\midrule
\endfirsthead
\toprule
Level & Retained structural information & Abstraction rule & Number of nodes \\

\midrule
\endhead
\bottomrule
\endfoot
\textbf{H1} & Ring-count category & Record the number of ring nodes generated by the approximate smallest-ring representation & 14 \\
\textbf{H2} & Connectivity topology between rings & Represent each ring as a node, retain connections between rings, and remove connection types & 498 \\
\textbf{H3} & Ring topology with connection types & Distinguish connection types between rings on the basis of H2 & 1,451 \\
\textbf{H4} & Minimal scaffold topology & Generalize atoms and bonds and reduce degree-2 nodes while retaining minimal ring structures & 14,378 \\
\textbf{H5} & Ring sizes and reduced connecting chains & Retain topology within rings and compress non-ring connecting chains of degree-2 nodes & 153,699 \\
\textbf{H6} & Complete connectivity graph of a generalized scaffold & Remove atom and bond types while retaining connectivity and connecting-chain lengths & 412,853 \\
\textbf{H7} & Exact Murcko scaffold with atom and bond types & Retain chemical types and canonicalize after removing stereochemistry, atom mapping, and isotope labels & 1,665,615 \\
\end{longtable}
\endgroup

\FloatBarrier
\subsection{Scaffold Navigation and Molecular Editing Actions}\label{sec:app-actions}

The MolHop harness provides four groups of structured actions, including scaffold navigation, molecular inspection, local editing, and search control, with a total of 9 external actions, as shown in Table \ref{tab:A2}. The agent selects subsequent actions based on structured information returned by the environment, including candidate categories, scaffold identifiers, editable sites, and action constraints. All actions are executed by the environment, which returns validated state feedback. The agent does not directly generate free-form SMILES to modify molecules.

Scaffold navigation actions query and select candidate structures in the hierarchical scaffold space. \texttt{nav\_where} returns the position of the current molecule in the H1-H7 hierarchical space and the regions available for exploration. \texttt{nav\_list} enumerates categories, subregions, and specific scaffolds at a specified level. \texttt{nav\_filter} applies structural constraints to H7 leaf nodes, including molecular weight, heavy-atom count, ring count, rotatable-bond count, heteroatom count, and elemental constraints. All filtering conditions apply to the same scaffold record and are evaluated jointly. The filtering results return a candidate set with stable identifiers for subsequent selection and further filtering. \texttt{nav\_pick} loads the target scaffold using a candidate identifier and uses it as the input structure for subsequent edit actions.

Molecular inspection actions identify editable positions in the current structure. \texttt{mol\_show} returns the corresponding site information for a specified action category, and its \texttt{all} mode returns counts of sites in different categories. \texttt{mol\_site} takes a target site identifier and returns the edit types supported at that position and information on available fragments.

The local editing action \texttt{mol\_edit} supports three types of editing actions for fragments, atoms and bonds, and ring structures, with a total of 9 subactions, as shown in Table \ref{tab:A3}. Each edit first undergoes environment validation, including RDKit molecular structure validity checks and a single-connected-component constraint check. Actions that pass validation generate a new molecular state and update the corresponding property feedback. Actions that fail validation return an error message, and the current state remains unchanged. Structural validity is ensured by the editing environment, and synthetic accessibility is subsequently evaluated using the SA score metric.

Search control tools manage episode states. \texttt{mol\_undo} restores historical states supported by the environment, and \texttt{mol\_done} submits the current candidate and ends the current search process.

\begingroup
\renewcommand{\thetable}{A2}
\begin{longtable}{@{}>{\raggedright\arraybackslash}p{\dimexpr 0.220\linewidth-2\tabcolsep\relax}>{\raggedright\arraybackslash}p{\dimexpr 0.200\linewidth-2\tabcolsep\relax}>{\raggedright\arraybackslash}p{\dimexpr 0.580\linewidth-2\tabcolsep\relax}@{}}
\caption{MolHop action space.}\label{tab:A2}\\
\toprule
Action group & \textbf{Action} & \textbf{Description} \\

\midrule
\endfirsthead
\toprule
Action group & \textbf{Action} & \textbf{Description} \\

\midrule
\endhead
\bottomrule
\endfoot
Scaffold navigation & \texttt{nav\_where} & Query the position of the current molecule in H1-H7 space and the scope available for exploration \\
Scaffold navigation & \texttt{nav\_list} & List categories, subregions, and specific scaffolds at a specified level or within a candidate set \\
Scaffold navigation & \texttt{nav\_filter} & Apply combined structural constraints to H7 leaf nodes and return a candidate set \\
Scaffold navigation & \texttt{nav\_pick} & Select a scaffold from previously returned candidates and load it as a new starting point for molecular editing \\
Molecular inspection & \texttt{mol\_show} & Enumerate editable sites for a specified action category, with \texttt{all} mode returning only the number of sites in each category \\
Molecular inspection & \texttt{mol\_site} & Obtain permitted actions based on site identifiers returned by \texttt{mol\_show}, and obtain the action and fragment identifiers allowed at that site \\
Molecular editing & \texttt{mol\_edit} & Execute a structural modification permitted by the environment and return the new structure and property feedback \\
Search control & \texttt{mol\_undo} & Undo an edit or restore a historical state supported by the environment \\
Search control & \texttt{mol\_done} & Submit the current molecule and end the episode \\
\end{longtable}
\endgroup

\begingroup
\renewcommand{\thetable}{A3}
\begin{longtable}{@{}>{\raggedright\arraybackslash}p{\dimexpr 0.160\linewidth-2\tabcolsep\relax}>{\raggedright\arraybackslash}p{\dimexpr 0.280\linewidth-2\tabcolsep\relax}>{\raggedright\arraybackslash}p{\dimexpr 0.560\linewidth-2\tabcolsep\relax}@{}}
\caption{Molecular editing actions implemented by mol\_edit.}\label{tab:A3}\\
\toprule
\textbf{Object} & \textbf{Action} & \textbf{Description} \\

\midrule
\endfirsthead
\toprule
\textbf{Object} & \textbf{Action} & \textbf{Description} \\

\midrule
\endhead
\bottomrule
\endfoot
Fragment & \texttt{add\_fragment} & Add a fragment at a permitted attachment site \\
Fragment & \texttt{delete\_fragment} & Delete a fragment whose removal is permitted \\
Fragment & \texttt{replace\_fragment} & Replace the current fragment with another fragment \\
Fragment & \texttt{fragment\_swap} & Replace a degree-2 atom whose two adjacent bonds are both non-ring bonds with a structural fragment having two attachment points to alter the connecting chain \\
Atoms and bonds & \texttt{mutate\_atom} & Modify the atom type at a permitted site \\
Atoms and bonds & \texttt{change\_bond} & Modify a permitted bond type or bond order \\
Ring structure & \texttt{open\_ring} & Open a ring through a permitted bond-breaking action \\
Ring structure & \texttt{close\_ring} & Form a bond between eligible sites to close a ring \\
Ring structure & \texttt{replace\_ring} & Replace a ring structure according to an available template \\
\end{longtable}
\endgroup

\FloatBarrier
\subsection{Trajectory Memory}\label{sec:app-memory}

MolHop memory is constructed deterministically by the harness from environment interaction to record the search state within a single episode. Memory updates are triggered by actual execution events, including scaffold loading, molecular editing, state restoration, and property evaluation. Whenever a tool response is generated, the harness reconstructs memory from the current environment state and returns it to the agent together with the tool result.

\begin{figure}[!htbp]
\centering
\renewcommand{\thefigure}{A2}
\includegraphics[width=\linewidth,height=0.70\textheight,keepaspectratio]{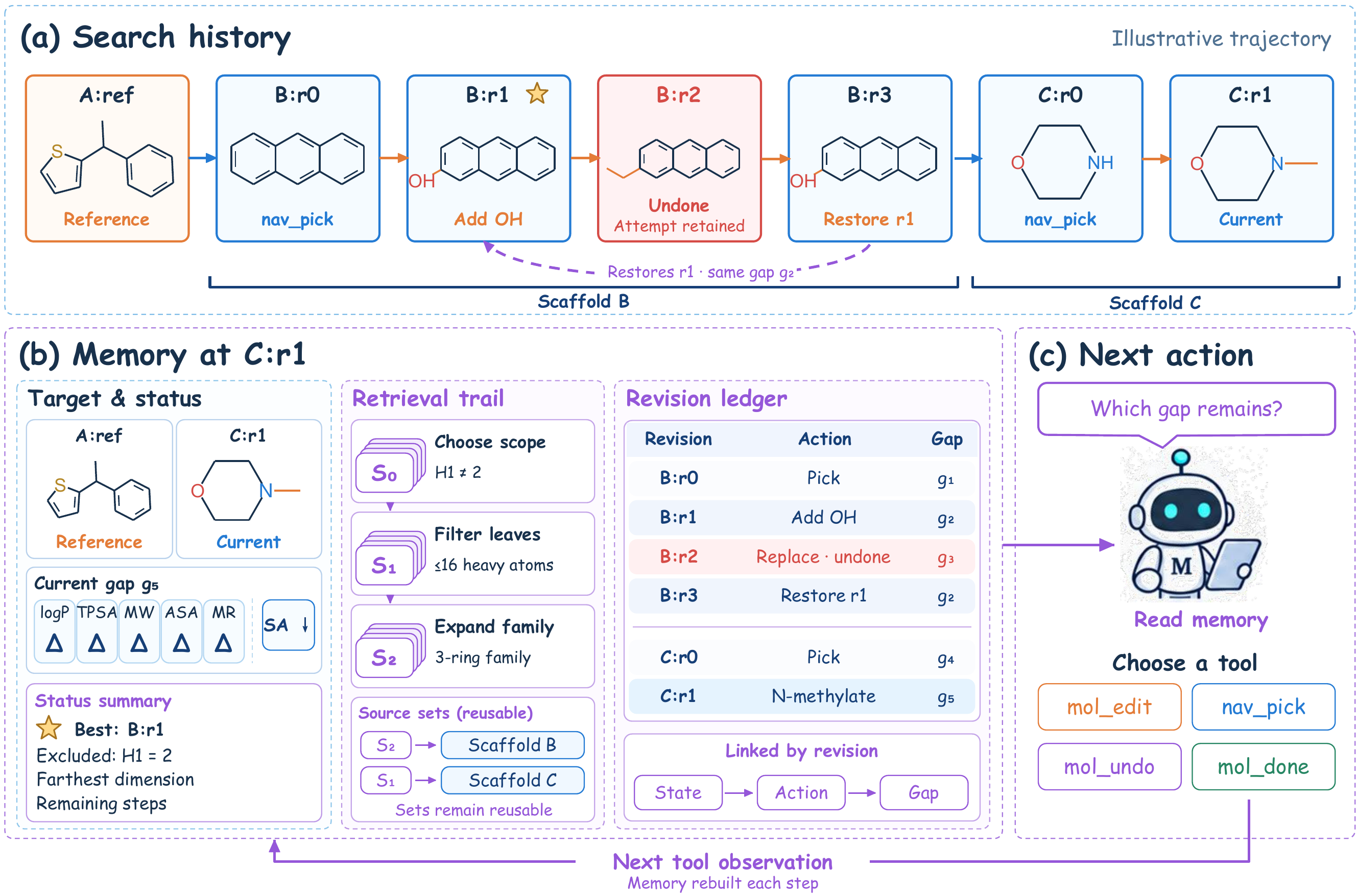}
\caption{Construction and use of trajectory memory in MolHop. (a) Search history within an illustrative trajectory, including scaffold transitions, molecular revisions, and undo operations. (b) Memory state at the current revision, summarizing target status, retrieval trail, revision ledger, and current property gaps. (c) The agent reads memory and selects the next action.}
\label{fig:A2}
\end{figure}

As shown in Figure \ref{fig:A2}, memory organizes the search process into a set of interrelated state representations, including the current target state, scaffold retrieval trail, revision ledger, and context for deciding the next action. This structure records changes in candidate molecules across different scaffolds and revisions and associates historical actions with the corresponding property gaps.

The information recorded in memory includes the reference target, current candidate state, property feedback, evaluation status, best historical record, scaffold visitation trajectory, editing trajectory, candidate-set state, and search budget, as shown in Table \ref{tab:A4}. Property feedback is directly bound to the corresponding molecular state, and structural identifiers distinguish different revisions to ensure consistency between historical evaluations and the current candidate state.

To accommodate a limited context window, the harness generates memory summaries according to predefined rules. By default, a state summary displays at most 10 segments of scaffold visitation records, retains the most recent 12 steps in each segment, and explicitly records the number of omitted entries. During inference, memory content in earlier tool responses can be replaced with summary placeholders, while the complete score log remains in the environment state. RL rollouts use a separate state compression process that updates candidate lists and historical entries before new observations enter the context.

All memory update and summarization rules are determined by the environment and do not depend on the LLM autonomously generating historical summaries. This design ensures that the same environment state corresponds to the same memory representation and makes memory a structured episode state that can be directly used for subsequent action selection.

\begingroup
\renewcommand{\thetable}{A4}
\begin{longtable}{@{}>{\raggedright\arraybackslash}p{\dimexpr 0.220\linewidth-2\tabcolsep\relax}>{\raggedright\arraybackslash}p{\dimexpr 0.390\linewidth-2\tabcolsep\relax}>{\raggedright\arraybackslash}p{\dimexpr 0.390\linewidth-2\tabcolsep\relax}@{}}
\caption{Trajectory Memory State Structure}\label{tab:A4}\\
\toprule
\textbf{Content} & \textbf{Recorded information} & \textbf{Role in search} \\

\midrule
\endfirsthead
\toprule
\textbf{Content} & \textbf{Recorded information} & \textbf{Role in search} \\

\midrule
\endhead
\bottomrule
\endfoot
\textbf{Reference target} & Starting molecule, five target properties, and initial SA score & Keep the optimization target fixed across multi-turn interactions \\
\textbf{Current property feedback} & Signed differences for each property, absolute relative deviations, match distance, normalized SA score, total cost, and the property with the largest deviation & Indicate which properties should be prioritized for adjustment in the next step and compare different candidates \\
\textbf{Evaluation status} & Current scaffold load identifier, molecular revision identifier, evaluated structure, and \texttt{gap\_is\_current} & Distinguish feedback on the current candidate from feedback on historical structures \\
\textbf{Best historical record} & Lowest cost observed so far, corresponding SMILES, step identifier, and revision identifier & Retain a search reference for review, which is not equivalent to the best candidate already satisfying joint \\
\textbf{Scaffold visitation trajectory} & Visited scaffold categories, abstraction levels, entry sources, and corresponding scaffolds & Record search transitions between different structural regions \\
\textbf{Within-category editing trajectory} & Revision identifiers, action summaries, property gaps, undo flags, and restoration targets & Associate local structural changes with property feedback \\
\textbf{Navigation and candidate-set state} & Current hierarchical position, established candidate sets, active candidate set, and a summary of filtering conditions & Support reuse of filtering results and switching of the search scope \\
\textbf{Search budget} & Turns used, remaining turns, number of tool steps, number of edits, and number of evaluations & Support allocation of the remaining budget among exploration, optimization, and submission \\
\end{longtable}
\endgroup

\FloatBarrier
\section{Experimental Setup}

\FloatBarrier
\subsection{Dataset Construction}\label{sec:app-dataset}

The experimental data are constructed from ChEMBL and the Enamine Screening Collection. All experiments use a unified offline scoring function to recalculate target properties and SA scores.

The drug molecule data are obtained from records of approved synthetic small molecules in the ChEMBL database. We select molecules with \texttt{max\_phase=4} and exclude entries marked as withdrawn. The Approved label in ChEMBL denotes the drug status recorded in the database and does not indicate that this study revalidates marketing status in different regions.

The commercial compound data are obtained from the August 2026 version of the Enamine Screening Collection. All molecules are first standardized using RDKit, including removal of records with nonempty SaltType labels and retention of the largest organic fragment in multicomponent molecules. Molecular parsing, connectivity checks, and metal-element filtering are then performed, followed by deduplication based on canonical SMILES.

For reference molecules from both data sources, we apply the same structural filtering criteria. The heavy-atom count is between 10 and 50, the ring count is between 1 and 6, the absolute value of Crippen logP is at least 0.5, and TPSA is at least 10 Å². The logP and TPSA conditions ensure stability in relative-error calculations and do not define a general range of drug-likeness. The five target properties and SA scores are all recalculated using the unified implementation within the project, rather than directly using property labels provided by the database or vendor.

The test sets are first deduplicated by Bemis--Murcko scaffold and then sampled in strata based on ring count. Within each stratum, representative molecules are selected using MaxMin diversity selection with Morgan fingerprints of radius 2 and length 2048 bits. All sampling procedures use the fixed random seed 42. For ChEMBL data, candidates sharing the same scaffold are selected by prioritizing first-in-class molecules, followed by records with earlier approval dates. Duplicate molecules between the two test sets are excluded using InChIKey.

Both SFT and RL training data exclude InChIKeys present in the test sets. SFT starting points additionally exclude the RL starting-point set. SFT and RL data sampling use random seeds 123 and 456, respectively, and the training--validation split uses random seed 42. Different molecules in the training sets may share a Murcko scaffold, but no starting molecules are identical to those in the test sets.

\begingroup
\renewcommand{\thetable}{B1}
\begin{longtable}{@{}>{\raggedright\arraybackslash}p{\dimexpr 0.250\linewidth-2\tabcolsep\relax}>{\raggedright\arraybackslash}p{\dimexpr 0.140\linewidth-2\tabcolsep\relax}>{\raggedright\arraybackslash}p{\dimexpr 0.140\linewidth-2\tabcolsep\relax}>{\raggedright\arraybackslash}p{\dimexpr 0.130\linewidth-2\tabcolsep\relax}>{\raggedright\arraybackslash}p{\dimexpr 0.340\linewidth-2\tabcolsep\relax}@{}}
\caption{Experimental Dataset Split}\label{tab:B1}\\
\toprule
Data use & ChEMBL & Enamine & Total size & Data split \\

\midrule
\endfirsthead
\toprule
Data use & ChEMBL & Enamine & Total size & Data split \\

\midrule
\endhead
\bottomrule
\endfoot
Drug test set & 200 & 0 & 200 & Scaffold deduplication, ring-count stratification, and diversity selection \\
Commercial test set & 0 & 200 & 200 & Scaffold deduplication, ring-count stratification, and diversity selection \\
SFT starting points & 500 & 1,500 & 2,000 & Stratified sampling, 1,800 train / 200 validation \\
RL starting points & 750 & 4,250 & 5,000 & Supplement with the remaining eligible samples, 4,872 train / 128 validation \\
\end{longtable}
\endgroup

\FloatBarrier
\subsection{Evaluation Details}

The property matching distance is calculated using absolute relative deviations. When a reference property is zero, the code retains an infinite relative error and does not alter the formal metric definition by adding a smoothing term.

The joint metric uses the total number of reference tasks as its denominator. Invalid outputs, missing outputs, and candidates that violate any constraint are counted as failures. Decomposed metrics such as H1 transition rate and SA improvement rate are used only to analyze the sources of failure under different constraints, and their statistical denominators are distinguished from that of joint.

The SA score is calculated using the unified project implementation. The normalized SA score is defined as:

\[SA_{norm}(x)=\frac{SA(x)}{SA(x_0)}\]

where a value below 1 indicates improvement relative to the starting point.

Structural similarity metrics are calculated based on Bemis--Murcko scaffolds. Scaffold Tanimoto uses Morgan fingerprints with a radius of 2 and a length of 2048 bits. Scaffold MCS Ratio is defined as the number of heavy atoms in the maximum common substructure divided by the number of heavy atoms in the smaller of the two scaffolds

\[MCS= \frac{atoms(MCS)} {\min(atoms(S_x),atoms(S_{x_0}))}\]

Search efficiency is defined as:
\vspace{-0.05cm}
\begin{equation*}
\mathrm{Efficiency}
=
\frac{\sum_i\mathrm{calls}_i}
{\sum_iJH_1(x_i,x_{0,i})}.
\end{equation*}
\FloatBarrier
\subsection{SFT and Agentic RL Training}\label{sec:app-training}

MolHop is trained using Qwen3.5 2B, 4B, and 9B models, yielding MolHop-2B, MolHop-4B, and MolHop-9B, respectively. We compare MolHop with a range of general-purpose and chemistry-oriented language models, including Llama-8B, Llama-70B\citep{grattafiori2024llama}, phi-4\citep{abdin2024phi}, Qwen3.5-2B, Qwen3.5-4B, Qwen3.5-9B, Qwen3.5-27B\citep{qwen2026qwen35}, Gemma4-31B\citep{team2026gemma}, ChemLLM-7B\citep{zhang2024chemllm}, ChemDFM-v2-14B\citep{zhao2025developing,zhao2025chemdfm}, DeepSeek-Flash-V4.1\citep{xu2026deepseek}, Grok-4.6, Gemini-3.1-Pro, GPT-5.6-Sol, and Claude-5-Opus. Training consists of supervised fine-tuning and reinforcement learning. Training consists of supervised fine-tuning and reinforcement learning. During supervised fine-tuning, Grok-4.6 generates demonstration trajectories within the MolHop harness. The demonstration data contain 2,000 dedicated starting points, use the H1 scaffold transition setting, and have a maximum interaction length of 50 turns. Each trajectory contains model actions, tool-call arguments, and environment feedback. The data include 1,214 trajectories with normal submission and 786 trajectories that reached the maximum interaction length. The training data are not filtered by final joint success outcomes, preserving different search behaviors. The data are split by starting point to ensure that trajectories generated from the same starting point do not appear in both the training and validation sets. SFT uses full-parameter fine-tuning, with the detailed configuration shown in Table \ref{tab:B2}.

Reinforcement learning is initialized from the corresponding SFT checkpoint and uses online GRPO training on a separate set of RL starting points. Multiple complete episodes are sampled for each starting point, and within-group advantages are calculated from terminal rewards. The GRPO configuration is shown in Table \ref{tab:B3}.

The three model sizes share the above training configuration. All training is performed on 4 NVIDIA A800 80GB GPUs, with the SFT and RL stages taking approximately 3 hours and 18 hours, respectively.

\begingroup
\renewcommand{\thetable}{B2}
\begin{longtable}{@{}>{\raggedright\arraybackslash}p{\dimexpr 0.500\linewidth-2\tabcolsep\relax}>{\raggedright\arraybackslash}p{\dimexpr 0.500\linewidth-2\tabcolsep\relax}@{}}
\caption{SFT training configuration}\label{tab:B2}\\
\toprule
\textbf{Setting} & \textbf{Configuration} \\

\midrule
\endfirsthead
\toprule
\textbf{Setting} & \textbf{Configuration} \\

\midrule
\endhead
\bottomrule
\endfoot
Number of training and validation trajectories & 1,800 and 200 \\
Fine-tuning method & Full-parameter fine-tuning, bf16, FSDP \\
Training epochs & 3 epochs, 675 optimization steps in total \\
Global batch size & 8 \\
Maximum sequence length & 16,384 tokens \\
Optimizer & AdamW \\
Learning rate & 0.00001 \\
Weight decay & 0.01 \\
Adam $\beta_1$ and $\beta_2$ & 0.9 and 0.999 \\
Validation and checkpoint saving & At the end of each epoch \\
\end{longtable}
\endgroup

\begingroup
\renewcommand{\thetable}{B3}
\begin{longtable}{@{}>{\raggedright\arraybackslash}p{\dimexpr 0.500\linewidth-2\tabcolsep\relax}>{\raggedright\arraybackslash}p{\dimexpr 0.500\linewidth-2\tabcolsep\relax}@{}}
\caption{GRPO training configuration}\label{tab:B3}\\
\toprule
\textbf{Setting} & \textbf{Configuration} \\

\midrule
\endfirsthead
\toprule
\textbf{Setting} & \textbf{Configuration} \\

\midrule
\endhead
\bottomrule
\endfoot
Number of training and validation starting points & 4,872 and 128 \\
Starting points per batch & 16 \\
Rollouts per starting point & 4 \\
Maximum assistant turns per episode & 50 \\
Training sampling temperature and top-p & 1.0 and 1.0 \\
Validation sampling temperature & 0 \\
Maximum prompt length & 4,096 tokens \\
Maximum response length & 32,768 tokens \\
Learning rate & 0.000001 \\
PPO minibatch size and update epochs & 16 and 1 \\
Policy clipping coefficient & 0.2 \\
KL loss and reward KL & Both disabled \\
Validation and checkpoint saving interval & Every 10 optimization steps \\
Checkpoint selection criterion & Joint on the separate RL validation set, with reward used to break ties \\
\end{longtable}
\endgroup

The three model sizes share the above training configuration.

The RL stage uses a terminal continuous cost

\[Cost(x, x_0) = MD(x, x_0) + SA(x) / SA(x_0)\]

When the agent completes \texttt{mol\_done} and obtains a valid terminal evaluation, the reward is defined as

\[Reward =max(-Cost,-4)\]

Trajectories without a valid submission or with an invalid terminal evaluation receive a reward of -5.

A successful submission denotes completion of the tool protocol and receipt of a terminal evaluation and does not imply satisfaction of joint@H1. The continuous reward provides a direction for property matching and SA optimization. Final performance is evaluated using joint@H1, which jointly requires structural transition, property matching, and SA improvement.

\FloatBarrier
\section{Additional Experimental Results}

\FloatBarrier
\subsection{Analysis of MolHop Harness on Frozen LLM Agents}\label{sec:app-frozen-agents}

To further analyze the sources of the joint@H1 improvement brought by the MolHop harness, we count candidates satisfying structural validity, property matching with MD\textless1, SA improvement, and H1 transition, as shown in Tables \ref{tab:C1} and \ref{tab:C2}. Each individual count represents the number of candidates satisfying the corresponding condition. The three-condition intersection represents the number of candidates simultaneously satisfying structural validity, property matching, and SA improvement, without adding the scaffold transition condition. The results show that, after integration with the harness, both models produce valid outputs for the full test-set size on both test sets, while H1 transition and property matching both improve substantially. For example, on the drug tasks, the H1 transition count of DeepSeek-Flash-V4.1 increases from 87 to 192, its MD\textless1 count increases from 107 to 183, and its joint success count increases from 31 to 105. The difference between the three-condition intersection and joint@H1 reflects the constraint imposed by scaffold transition on the final success rate. This decomposition indicates that the harness improves property optimization and also helps the agent transition between structural regions.

\begingroup
\renewcommand{\thetable}{C1}
\begin{longtable}{@{}>{\raggedright\arraybackslash}p{\dimexpr 0.300\linewidth-2\tabcolsep\relax}>{\raggedright\arraybackslash}p{\dimexpr 0.100\linewidth-2\tabcolsep\relax}>{\raggedright\arraybackslash}p{\dimexpr 0.100\linewidth-2\tabcolsep\relax}>{\raggedright\arraybackslash}p{\dimexpr 0.120\linewidth-2\tabcolsep\relax}>{\raggedright\arraybackslash}p{\dimexpr 0.100\linewidth-2\tabcolsep\relax}>{\raggedright\arraybackslash}p{\dimexpr 0.100\linewidth-2\tabcolsep\relax}>{\raggedright\arraybackslash}p{\dimexpr 0.180\linewidth-2\tabcolsep\relax}@{}}
\caption{Drug dataset}\label{tab:C1}\\
\toprule
 & joint & Valid & md<1 & SA$\uparrow$ & H1 & Valid$\land$md$\land$SA \\

\midrule
\endfirsthead

\toprule
 & joint & Valid & md<1 & SA$\uparrow$ & H1 & Valid$\land$md$\land$SA \\

\midrule
\endhead

\bottomrule
\endfoot

Grok  & 82 & 170 & 148 & 144 & 103 & 123 \\

\rowcolor{gray!15}
Grok + harness & \textbf{128} & \textbf{200} & \textbf{181} & \textbf{146} & \textbf{190} & \textbf{134} \\

DeepSeek  & 31 & 187 & 107 & 142 & 87 & 70 \\

\rowcolor{gray!15}
DeepSeek + harness & \textbf{105} & \textbf{200} & \textbf{183} & \textbf{123} & \textbf{192} & \textbf{111} \\

\end{longtable}
\endgroup

\begingroup
\renewcommand{\thetable}{C2}
\begin{longtable}{@{}>{\raggedright\arraybackslash}p{\dimexpr 0.300\linewidth-2\tabcolsep\relax}>{\raggedright\arraybackslash}p{\dimexpr 0.100\linewidth-2\tabcolsep\relax}>{\raggedright\arraybackslash}p{\dimexpr 0.100\linewidth-2\tabcolsep\relax}>{\raggedright\arraybackslash}p{\dimexpr 0.120\linewidth-2\tabcolsep\relax}>{\raggedright\arraybackslash}p{\dimexpr 0.100\linewidth-2\tabcolsep\relax}>{\raggedright\arraybackslash}p{\dimexpr 0.100\linewidth-2\tabcolsep\relax}>{\raggedright\arraybackslash}p{\dimexpr 0.180\linewidth-2\tabcolsep\relax}@{}}
\caption{Commercial dataset}\label{tab:C2}\\
\toprule
 & joint & Valid & md<1 & SA$\uparrow$ & H1 & Valid$\land$md$\land$SA \\

\midrule
\endfirsthead

\toprule
 & joint & Valid & md<1 & SA$\uparrow$ & H1 & Valid$\land$md$\land$SA \\

\midrule
\endhead

\bottomrule
\endfoot

Grok  & 73 & 166 & 145 & 137 & 86 & 119 \\

\rowcolor{gray!15}
Grok + harness & \textbf{138} & \textbf{200} & \textbf{186} & \textbf{149} & \textbf{195} & \textbf{140} \\

DeepSeek  & 35 & 186 & 128 & 134 & 82 & 85 \\

\rowcolor{gray!15}
DeepSeek + harness & \textbf{121} & \textbf{200} & \textbf{190} & \textbf{136} & \textbf{188} & \textbf{129} \\

\end{longtable}
\endgroup

\FloatBarrier
\subsection{Detailed Agentic RL Results}

To supplement the overall comparison in the main text, we further count the outputs of different models satisfying structural validity, property matching, SA improvement, and H1 transition, as shown in Tables \ref{tab:C3} and \ref{tab:C4}. These decomposed results show the constituent conditions of joint@H1. Some generation models can generate valid candidates with similar properties, but some candidates do not satisfy the H1 transition condition and are therefore excluded from joint@H1. For example, some strong generation models satisfy the three conditions of validity, MD\textless1, and SA improvement but still remain within the reference scaffold family.

\begin{figure}[!htbp]
\centering
\renewcommand{\thefigure}{C2}
\includegraphics[width=\linewidth,height=0.70\textheight,keepaspectratio]{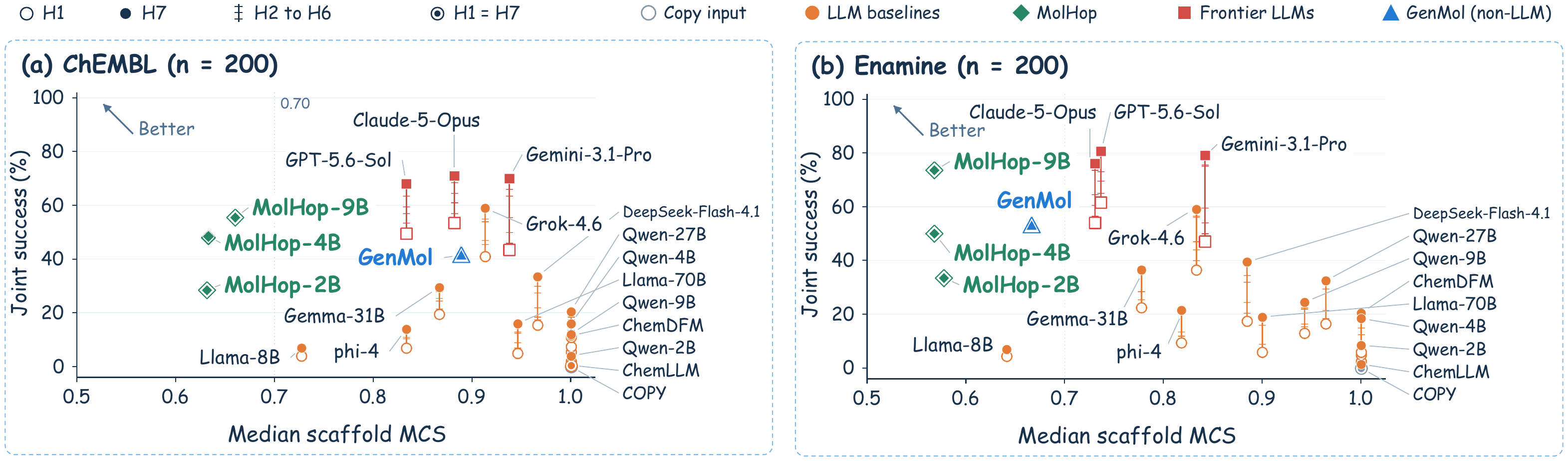}
\caption{Joint success versus median scaffold MCS under different scaffold transition criteria from H1 to H7. Higher joint success and lower MCS correspond to the upper-left region. LLM-based baselines improve mainly under relaxed scaffold criteria, while MolHop and GenMol remain stable.}
\label{fig:C2}
\end{figure}

We further evaluate joint success under different scaffold transition criteria from H1 to H7. As shown in Figure \ref{fig:C2}, MolHop and GenMol remain largely unchanged because their objectives already include H1 scaffold transition. In contrast, LLM-based generation baselines show higher joint success under relaxed scaffold criteria, resulting in an upward shift of points. However, this improvement mainly reflects easier scaffold constraints rather than stronger scaffold hopping ability, as scaffold MCS does not substantially decrease. Most baseline successes therefore still correspond to molecular optimization rather than cross-scaffold discovery.

\begin{figure}[!htbp]
\centering
\renewcommand{\thefigure}{C1}
\includegraphics[width=\linewidth,height=0.70\textheight,keepaspectratio]{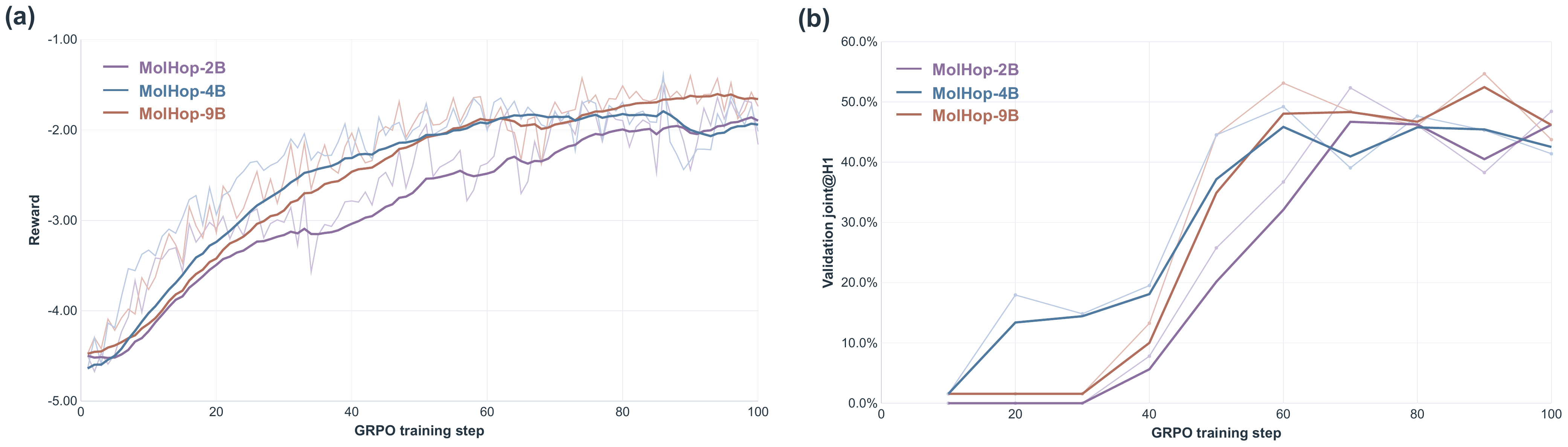}
\caption{Agentic RL training and cross-scaffold discovery performance of MolHop agents. (a) GRPO training reward curves for MolHop-2B, MolHop-4B, and MolHop-9B. (b) Validation joint@H1 performance during GRPO training. }
\label{fig:C1}
\end{figure}

\begingroup
\renewcommand{\thetable}{C3}
\begin{longtable}{@{}>{\raggedright\arraybackslash}p{\dimexpr 0.300\linewidth-2\tabcolsep\relax}>{\raggedright\arraybackslash}p{\dimexpr 0.100\linewidth-2\tabcolsep\relax}>{\raggedright\arraybackslash}p{\dimexpr 0.100\linewidth-2\tabcolsep\relax}>{\raggedright\arraybackslash}p{\dimexpr 0.120\linewidth-2\tabcolsep\relax}>{\raggedright\arraybackslash}p{\dimexpr 0.100\linewidth-2\tabcolsep\relax}>{\raggedright\arraybackslash}p{\dimexpr 0.100\linewidth-2\tabcolsep\relax}>{\raggedright\arraybackslash}p{\dimexpr 0.180\linewidth-2\tabcolsep\relax}@{}}
\caption{Drug dataset}\label{tab:C3}\\
\toprule
 & joint & Valid & md<1 & SA$\uparrow$ & H1 & Valid$\land$md$\land$SA \\

\midrule
\endfirsthead
\toprule
 & joint & Valid & md<1 & SA$\uparrow$ & H1 & Valid$\land$md$\land$SA \\

\midrule
\endhead
\bottomrule
\endfoot
MolHop-2B & 57 & 200 & 88 & 134 & 199 & 57 \\
MolHop-4B & 96 & 200 & 137 & 139 & 197 & 97 \\
MolHop-9B & 111 & 200 & 142 & 157 & 195 & 111 \\
Gemini-3.1-Pro & 87 & 199 & 186 & 154 & 109 & 144 \\
GPT-5.6-sol & 99 & 196 & 166 & 168 & 132 & 141 \\
Claude-5-Opus & 107 & 192 & 162 & 173 & 132 & 150 \\
\end{longtable}
\endgroup

\begingroup
\renewcommand{\thetable}{C4}
\begin{longtable}{@{}>{\raggedright\arraybackslash}p{\dimexpr 0.300\linewidth-2\tabcolsep\relax}>{\raggedright\arraybackslash}p{\dimexpr 0.100\linewidth-2\tabcolsep\relax}>{\raggedright\arraybackslash}p{\dimexpr 0.100\linewidth-2\tabcolsep\relax}>{\raggedright\arraybackslash}p{\dimexpr 0.120\linewidth-2\tabcolsep\relax}>{\raggedright\arraybackslash}p{\dimexpr 0.100\linewidth-2\tabcolsep\relax}>{\raggedright\arraybackslash}p{\dimexpr 0.100\linewidth-2\tabcolsep\relax}>{\raggedright\arraybackslash}p{\dimexpr 0.180\linewidth-2\tabcolsep\relax}@{}}
\caption{Commercial dataset}\label{tab:C4}\\
\toprule
 & joint & Valid & md<1 & SA$\uparrow$ & H1 & Valid$\land$md$\land$SA \\

\midrule
\endfirsthead
\toprule
 & joint & Valid & md<1 & SA$\uparrow$ & H1 & Valid$\land$md$\land$SA \\

\midrule
\endhead
\bottomrule
\endfoot
MolHop-2B & 67 & 200 & 100 & 136 & 200 & 67 \\
MolHop-4B & 105 & 200 & 144 & 137 & 197 & 105 \\
MolHop-9B & 147 & 200 & 165 & 170 & 200 & 147 \\
Gemini-3.1-Pro & 94 & 197 & 190 & 165 & 105 & 159 \\
GPT-5.6-sol & 123 & 199 & 180 & 182 & 143 & 163 \\
Claude-5-Opus & 108 & 195 & 159 & 189 & 132 & 154 \\
\end{longtable}
\endgroup

\FloatBarrier
\subsection{Scaffold Library Ablation Analysis}\label{sec:app-scaffold-ablation}

To analyze why reward and joint@H1 are inconsistent after removing the scaffold library, we diagnose the terminal candidates generated in the later stage of training by the model without the scaffold library. Specifically, we examine the final output states of 400 evaluation samples at step 150, including satisfaction of the property matching, SA improvement, and scaffold transition conditions.

The results show that this variant can effectively optimize the property matching objective, with 96.1\% of candidates satisfying MD\textless1. However, terminal SA scores do not improve correspondingly, with a median SA score change of +0.393 relative to the starting point. Further analysis shows that 67.7\% of candidates already satisfy the three conditions of structural validity, property matching, and H1 transition but are excluded from joint@H1 because their SA scores do not meet the requirement for strict improvement. The final joint success rate is 19.4\%.

These results indicate that, after removing the scaffold library, the search process relies more on molecular editing of the current structure for optimization. This strategy has two limitations. The absence of explicit scaffold navigation reduces the model's ability to enter distant structural regions, and some candidates cannot consistently complete the required scaffold transition. In addition, although local editing can reduce property distance, it struggles to ensure simultaneous SA improvement. Consequently, the model tends to obtain candidates with better property matching but cannot consistently coordinate the three objectives of structural transition, property matching, and synthetic accessibility.

\FloatBarrier
\subsection{Harness Tool Use and Tool-Call Efficiency}\label{sec:app-tool-efficiency}

The MolHop harness provides the agent with structured scaffold navigation, molecular editing, and search control interfaces. We further analyze the behavioral patterns and search costs of different models during tool use.

\textbf{As shown in Table \ref{tab:C5}}, the MolHop models have lower search costs on both test sets. Compared with general LLM agents, MolHop agents typically use fewer tool calls to obtain a joint@H1-successful candidate. MolHop models of different sizes maintain high Tool Execution Success Rate values, denoted TESR, indicating that the efficiency improvement comes from more effective search decisions rather than an increase in invalid tool calls.

\textbf{Figure \ref{fig:3}} shows the trajectory distributions of different models across four action groups, Navigation, Scaffold Selection, Edit, and Finish. Overall, MolHop agents and general LLM agents exhibit different tool-use patterns. MolHop models can allocate the interaction budget among scaffold exploration, candidate selection, and structural editing based on the search state, whereas general LLM agents typically continue editing-based exploration over longer trajectories. MolHop-2B is more inclined to retrieve candidates satisfying the conditions from the scaffold space and submit early, which explains its shorter average interaction length. Larger MolHop agents incorporate more edit actions to further optimize properties.

Overall, the policy learned by MolHop can use the structured action space to organize the search process and reduce redundant exploration while maintaining the structural transition and property optimization constraints.

\begingroup
\small
\renewcommand{\thetable}{C5}
\begin{longtable}{@{}>{\raggedright\arraybackslash}p{\dimexpr 0.260\linewidth-2\tabcolsep\relax}>{\raggedright\arraybackslash}p{\dimexpr 0.110\linewidth-2\tabcolsep\relax}>{\raggedright\arraybackslash}p{\dimexpr 0.135\linewidth-2\tabcolsep\relax}>{\raggedright\arraybackslash}p{\dimexpr 0.125\linewidth-2\tabcolsep\relax}>{\raggedright\arraybackslash}p{\dimexpr 0.110\linewidth-2\tabcolsep\relax}>{\raggedright\arraybackslash}p{\dimexpr 0.135\linewidth-2\tabcolsep\relax}>{\raggedright\arraybackslash}p{\dimexpr 0.125\linewidth-2\tabcolsep\relax}@{}}
\caption{MolHop Tool-Use Efficiency Comparison}\label{tab:C5}\\
\toprule
 & \multicolumn{3}{c}{\textbf{Drug molecules}} & \multicolumn{3}{c}{\textbf{Commercial molecules}} \\\cmidrule(lr){2-4}\cmidrule(lr){5-7}
Model & TESR\textsuperscript{a} & Efficiency$\downarrow$ & Avg Turns & TESR\textsuperscript{a} & Efficiency$\downarrow$ & Avg Turns \\

\midrule
\endfirsthead

\toprule
 & \multicolumn{3}{c}{\textbf{Drug molecules}} & \multicolumn{3}{c}{\textbf{Commercial molecules}} \\\cmidrule(lr){2-4}\cmidrule(lr){5-7}
Model & TESR\textsuperscript{a} & Efficiency$\downarrow$ & Avg Turns & TESR\textsuperscript{a} & Efficiency$\downarrow$ & Avg Turns \\

\midrule
\endhead

\bottomrule
\endfoot

\rowcolor{gray!15}
MolHop-9B & \textbf{86.2\%} & \textbf{28.0} & \textbf{15.6} & \textbf{87.8\%} & \textbf{20.0} & \textbf{14.7} \\

\rowcolor{gray!15}
MolHop-4B & \textbf{74.2\%} & \textbf{58.2} & \textbf{27.9} & \textbf{72.8\%} & \textbf{55.9} & \textbf{27.9} \\

\rowcolor{gray!15}
MolHop-2B & \textbf{79.7\%} & \textbf{25.7} & \textbf{7.3} & \textbf{80.8\%} & \textbf{21.1} & \textbf{7.1} \\

DeepSeek-Flash-V4.1 & 94.0\% & 90.1 & 45.9 & 94.3\% & 79.5 & 46.6 \\
Grok-4.6 & 78.4\% & 76.0 & 48.6 & 78.3\% & 70.0 & 48.4 \\

\multicolumn{7}{@{}l}{\footnotesize\textsuperscript{a} TESR denotes tool execution success rate.}\\

\end{longtable}
\endgroup

\FloatBarrier
\section{OOD Property Optimization beyond the Restricted Visible Scaffold Library}\label{sec:app-ood}

We further examine a stricter scenario in which the property distribution of target molecules lies outside the central region of the visible scaffold library, assessing whether an LLM integrated with the MolHop harness can still obtain candidates satisfying the constraints through editing. This experiment distinguishes the ability to directly retrieve suitable scaffolds from the scaffold library from the ability to continue property optimization within the harness.

To construct this scenario, we define a property visibility mask, \texttt{clip\_mask}, based on the complete scaffold library. Specifically, we calculate the 25th and 75th percentiles of each of the five properties logP, TPSA, MW, LabuteASA, and MolMR for molecules in the scaffold library and retain only H7 leaf nodes whose five properties all fall within the corresponding {[}p25,p75{]} intervals. The resulting thresholds are shown in Table \ref{tab:D1}. After applying this mask, 287,888 leaf nodes remain in the scaffold library, accounting for 17.3\% of all leaf nodes.

We then select a set of OOD targets from the 400 reference molecules in the ChEMBL-200 and Enamine-200 test sets. All five properties of each reference molecule fall within the above {[}p25,p75{]} box, but its nearest-neighbor match distance in the restricted scaffold library satisfies \(md\geq1\). This yields 83 targets, including 58 from ChEMBL and 25 from Enamine. For each target, we retrieve the leaf node with the smallest match distance in the restricted scaffold library and treat this result as the scaffold library upper bound. Because the target definition excludes close neighbors within the library, the md values of all 83 upper-bound candidates are at least 1.009, their mean md is 2.376, and their joint@H1 values are all 0.

We then integrate DeepSeek-Flash-V4.1 with the MolHop harness to search the same restricted scaffold library, with a maximum interaction budget of 100 turns. The results are shown in Table \ref{tab:D2}. Compared with scaffold library retrieval alone, DeepSeek + harness achieves a joint@H1 of 43.4\%, or 36/83, on the 83 OOD targets. Of the candidates, 90.4\%, or 75/83, satisfy md\textless1, and the mean match distance decreases from 2.376 to 0.542. This result indicates that, when the visible scaffold library itself cannot provide candidates close to the targets, the agent can use the editing actions provided by the harness to further optimize properties on the basis of selected scaffolds and recover some joint successes.

Figure \ref{fig:D1} shows a two-dimensional PCA projection of the target molecules, restricted scaffold library samples, and DeepSeek + harness outputs based on the five properties. The restricted scaffold library samples cluster in a smaller central region, while the 83 targets are widely distributed outside it. DeepSeek outputs move from the visible region of the library toward the regions containing the targets. Both jointly successful and unsuccessful samples show a clear tendency to move away from the library center, indicating that the task requires the model to continue editing-based optimization beyond the property range covered by bare scaffold retrieval.

% \begin{figure}[!htbp]
% \centering
% \renewcommand{\thefigure}{5}
% \includegraphics[width=0.75\linewidth,height=0.70\textheight,keepaspectratio]{figures/image5.pdf}
% \caption{OOD targets, restricted-library samples, and DeepSeek outputs in the property PCA space. PCA projection based on five molecular properties, logP, TPSA, MW, Labute ASA, and MolMR. Red points denote OOD targets, light blue points denote restricted scaffold library samples, and green points denote DeepSeek-Flash-V4.1 + harness outputs.}
% \label{fig:D1}
% \end{figure}

\begingroup
\renewcommand{\thetable}{D1}
\begin{longtable}{@{}>{\raggedright\arraybackslash}p{\dimexpr 0.400\linewidth-2\tabcolsep\relax}>{\raggedright\arraybackslash}p{\dimexpr 0.300\linewidth-2\tabcolsep\relax}>{\raggedright\arraybackslash}p{\dimexpr 0.300\linewidth-2\tabcolsep\relax}@{}}
\caption{Property Range Statistics for Restricted Scaffold Library}\label{tab:D1}\\
\toprule
Property & p25 & p75 \\

\midrule
\endfirsthead
\toprule
Property & p25 & p75 \\

\midrule
\endhead
\bottomrule
\endfoot
logP & 1.50 & 3.26 \\
TPSA & 52.83 & 84.50 \\
MW & 274.32 & 361.43 \\
LabuteASA & 116.73 & 153.55 \\
MolMR & 75.13 & 100.20 \\
\end{longtable}
\endgroup

\begingroup
\renewcommand{\thetable}{D2}
\begin{longtable}{@{}>{\raggedright\arraybackslash}p{\dimexpr 0.320\linewidth-2\tabcolsep\relax}>{\raggedright\arraybackslash}p{\dimexpr 0.250\linewidth-2\tabcolsep\relax}>{\raggedright\arraybackslash}p{\dimexpr 0.250\linewidth-2\tabcolsep\relax}>{\raggedright\arraybackslash}p{\dimexpr 0.180\linewidth-2\tabcolsep\relax}@{}}
\caption{OOD Property Optimization Results}\label{tab:D2}\\
\toprule
Method & joint@H1 & md<1 & Mean md $\downarrow$ \\

\midrule
\endfirsthead

\toprule
Method & joint@H1 & md<1 & Mean md $\downarrow$ \\

\midrule
\endhead

\bottomrule
\endfoot

Scaffold library upper bound & 0.0\% (0/83) & 0.0\% (0/83) & 2.376 \\

\rowcolor{gray!15}
DeepSeek + harness & \textbf{43.4\% (36/83)} & \textbf{90.4\% (75/83)} & \textbf{0.542} \\

\end{longtable}
\endgroup

\FloatBarrier
\section{Examples}
\begin{figure}[!htbp]
\centering
\renewcommand{\thefigure}{E1}
\includegraphics[width=\linewidth,height=0.70\textheight,keepaspectratio]{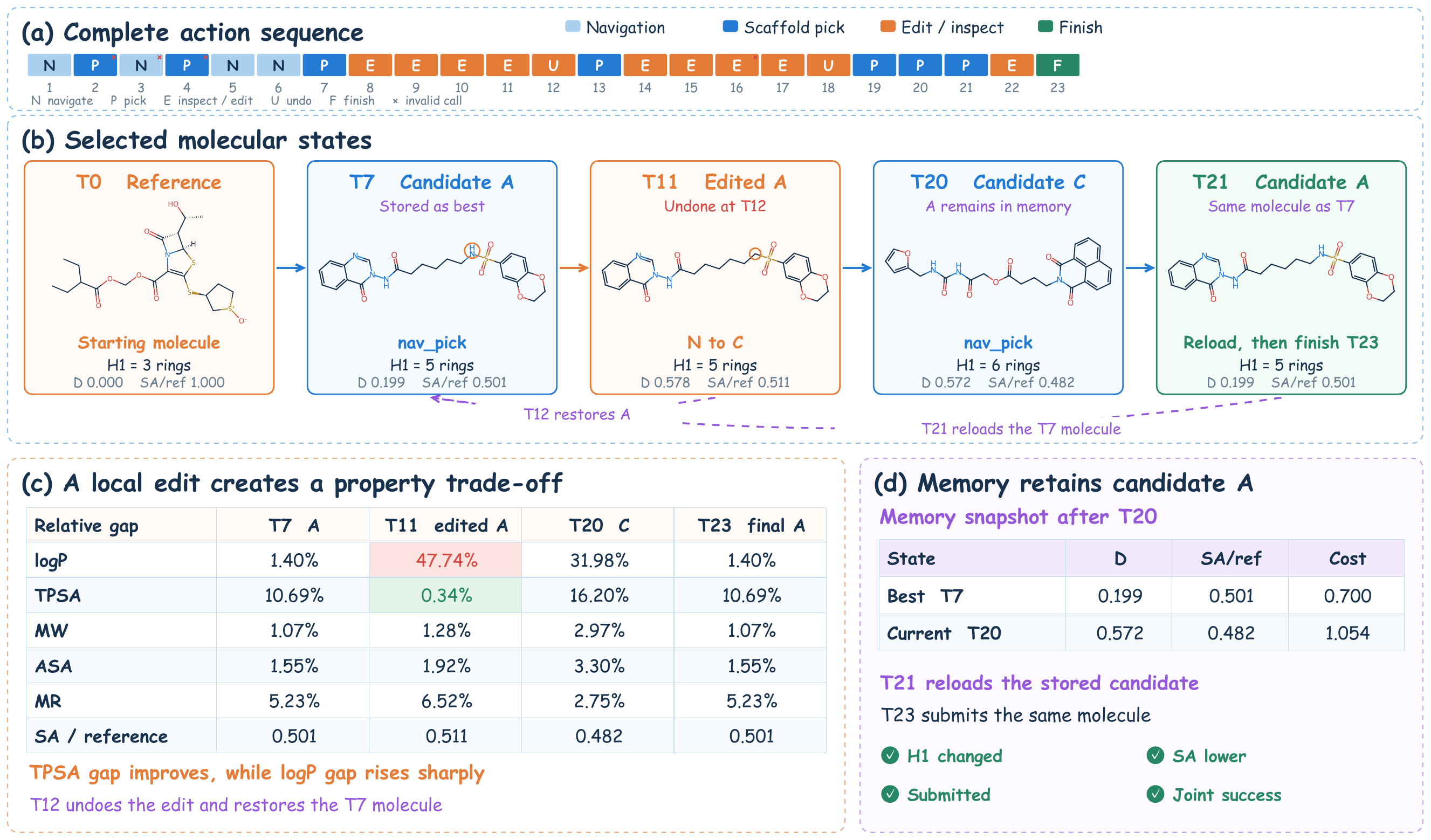}
\caption{Recovery of a candidate retained in memory.
In this successful MolHop-4B trajectory, an N-to-C edit improves TPSA matching but worsens logP matching and is undone. After exploring other scaffolds, the agent reloads candidate A at turn 21 and submits it at turn 23, achieving joint success.}
\label{fig:E1}
\end{figure}

\begin{figure}[!htbp]
\centering
\renewcommand{\thefigure}{E2}
\includegraphics[width=\linewidth,height=0.70\textheight,keepaspectratio]{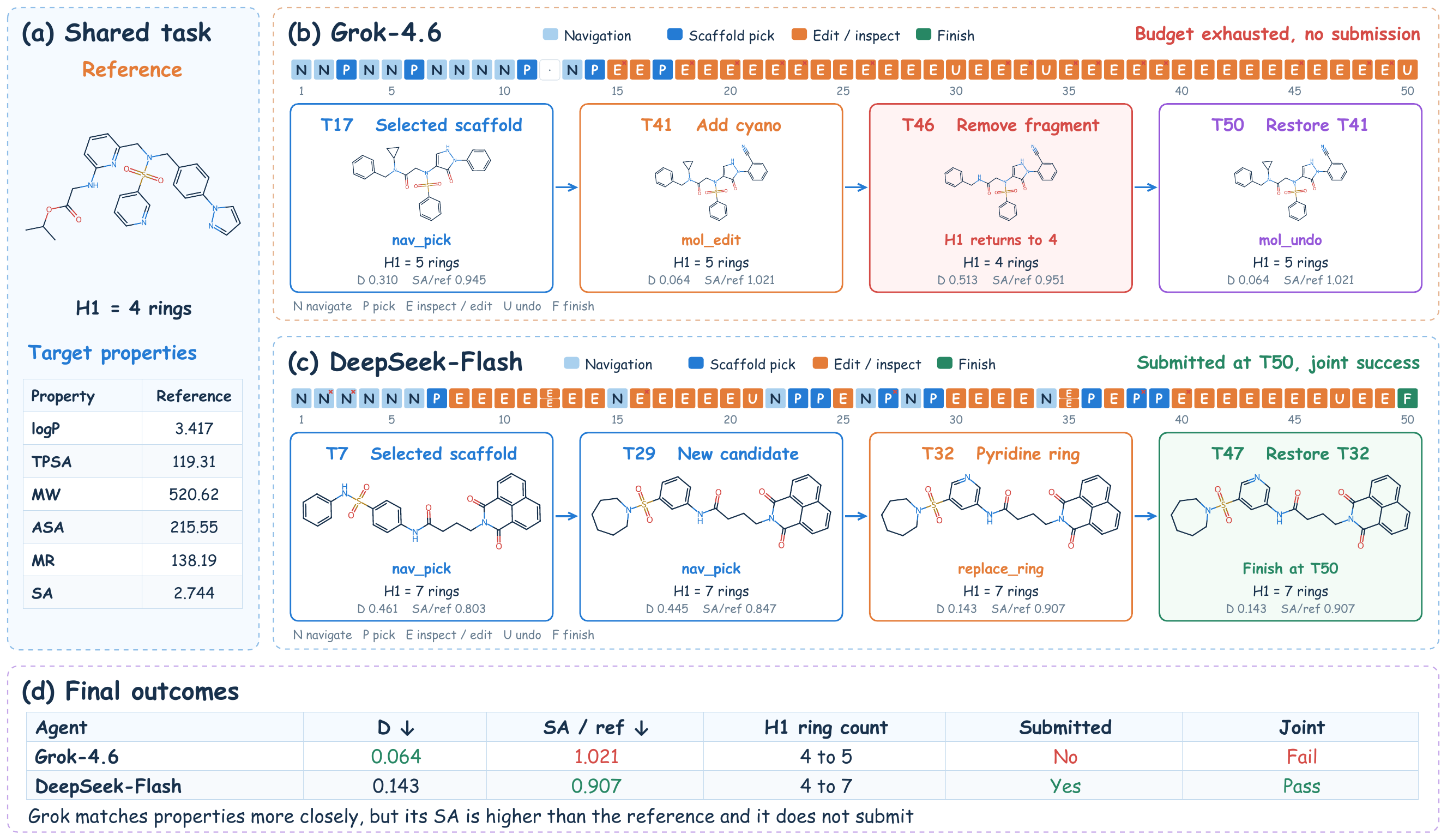}
\caption{Comparison from the same starting molecule.
Under identical 50-turn budgets, Grok-4.6 attains a smaller property distance but ends with a higher SA than the reference and no submission. DeepSeek-Flash-V4.1 submits a candidate with a lower SA and achieves joint success.}
\label{fig:E2}
\end{figure}

\begin{figure}[!htbp]
\centering
\renewcommand{\thefigure}{E3}
\includegraphics[width=\linewidth,height=0.70\textheight,keepaspectratio]{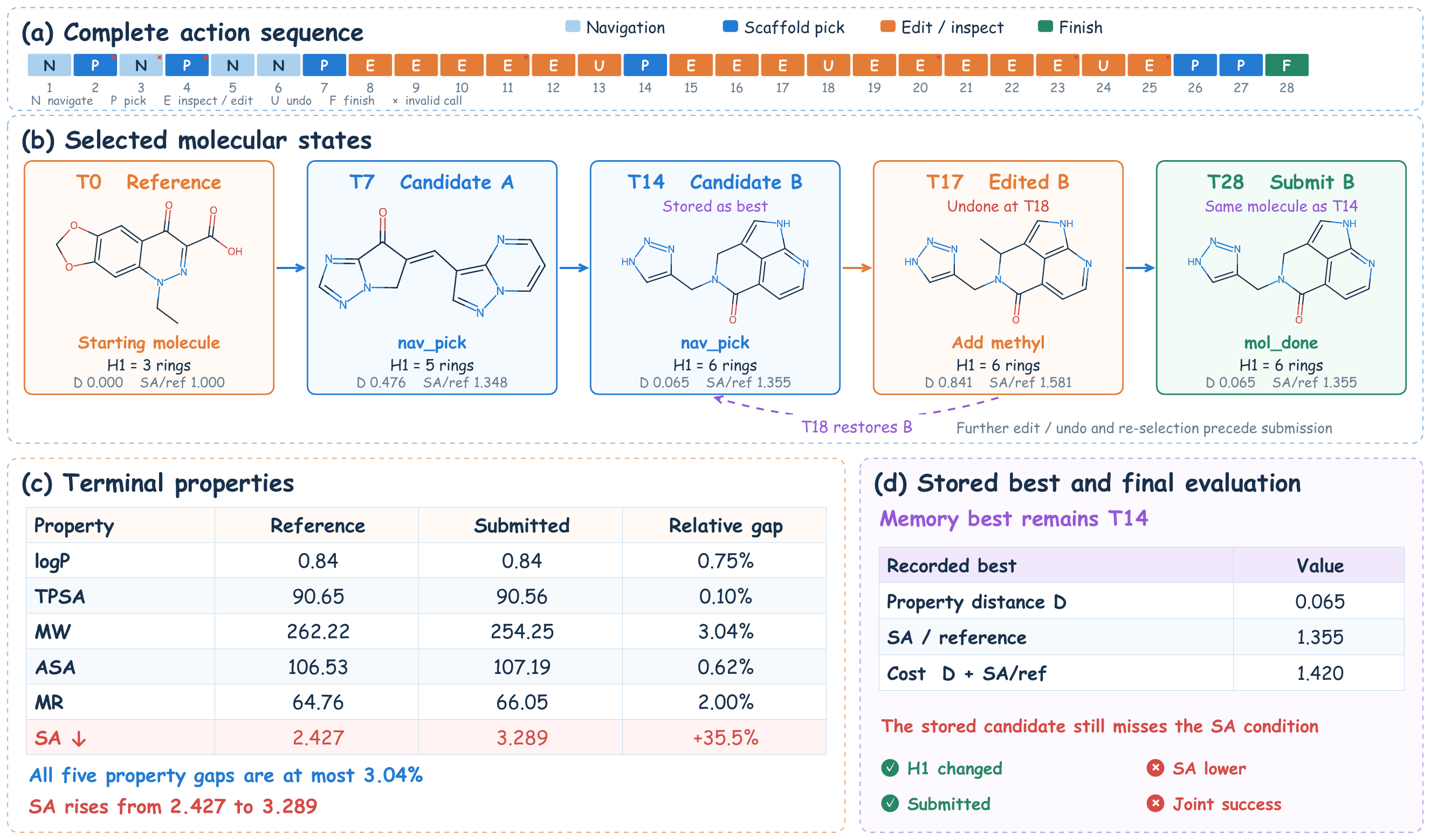}
\caption{MolHop-4B failure despite close property matching.
The agent submits its recorded best candidate at turn 28. All five relative property gaps are at most 3.04\%, but SA increases from 2.427 to 3.289. The submitted candidate therefore fails the SA condition and the joint objective.}
\label{fig:E3}
\end{figure}

\begin{figure}[!htbp]
\centering
\renewcommand{\thefigure}{E4}
\includegraphics[width=\linewidth,height=0.70\textheight,keepaspectratio]{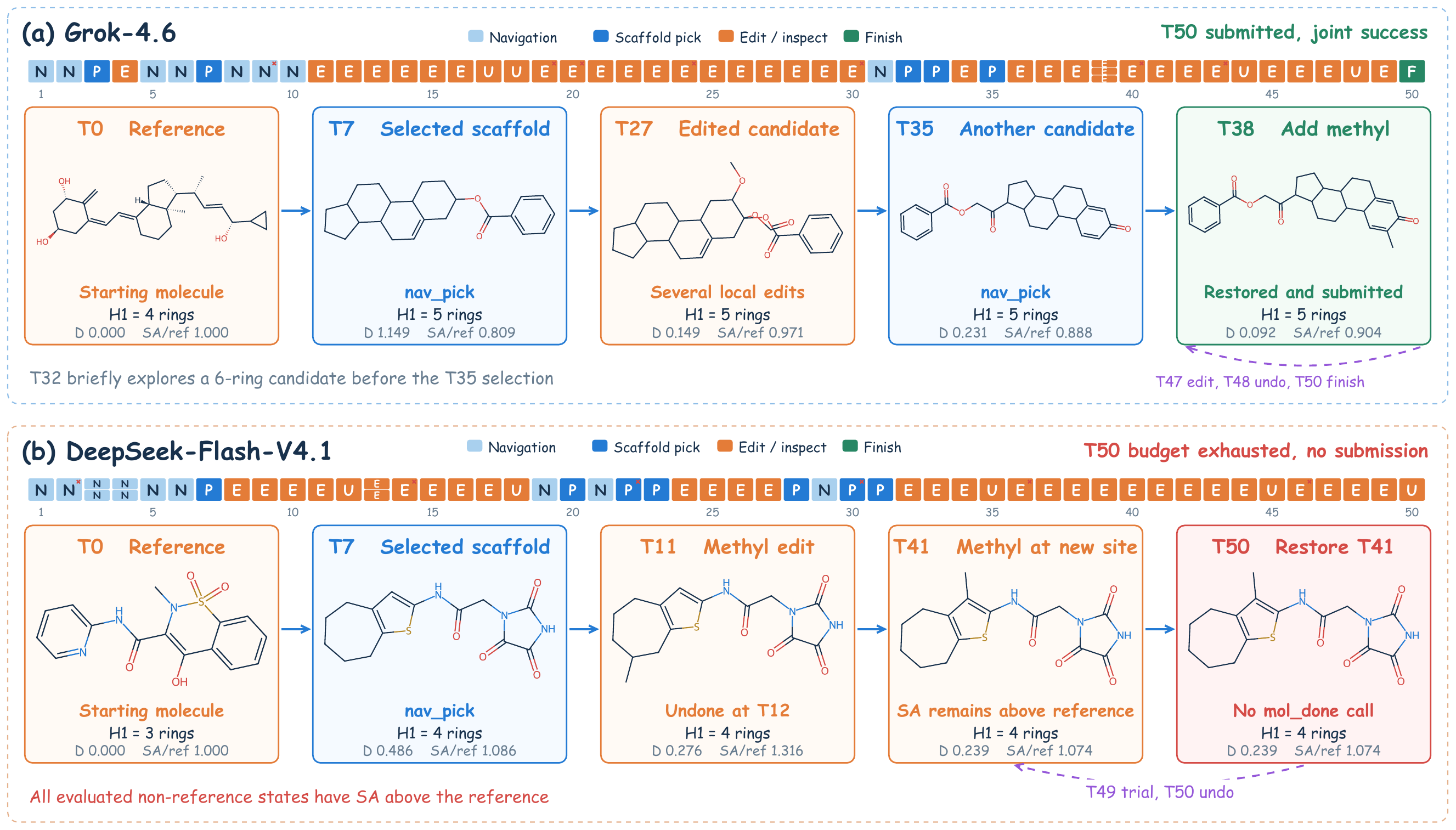}
\caption{Additional baseline trajectories from different starting molecules.
Grok-4.6 explores alternative scaffolds and local edits, restores a previous candidate, and submits successfully at turn 50. DeepSeek-Flash-V4.1 exhausts its budget after an undo operation without submitting. All of its evaluated non-reference states have SA scores above the reference.}
\label{fig:E4}
\end{figure}

Six recorded ChEMBL trajectories illustrate how scaffold navigation, local editing, and recovery of previously evaluated candidates interact under the joint@H1 objective. We show the complete tool-use sequence and selected molecular states for each episode. A successful MolHop-4B case highlights a local property trade-off and a subsequent return to a candidate retained in memory (Figure \ref{fig:E1}). A comparison from the same starting molecule shows why closer property matching alone does not ensure joint success (Figure \ref{fig:E2}). A MolHop-4B failure shows an unmet SA condition despite accurate matching of all five target properties (Figure \ref{fig:E3}). Two further baseline examples show successful submission and budget exhaustion on different tasks (Figure \ref{fig:E4}). Throughout, D denotes the sum of five normalized absolute property deviations, and SA/ref is the candidate SA score divided by the reference SA score. Lower values are preferred for both.

\end{document}